\documentclass[journal]{IEEEtran}
\usepackage{booktabs}
\usepackage{algorithmic}
\usepackage{graphicx}
\usepackage{textcomp}
\usepackage{booktabs}
\usepackage{multirow}
\usepackage{hhline}
\usepackage{float}
\usepackage{xcolor}
\usepackage{hyperref}
\usepackage{authblk}
\usepackage{amsmath,amssymb,amsfonts}
\usepackage[authoryear]{natbib}
\usepackage{caption}
\usepackage{subcaption}
\begin{document}

\title{Benchmarking and scaling of deep learning models for land cover image classification}

\author[1]{Ioannis Papoutsis}
\author[12]{Nikolaos Ioannis Bountos}
\author[1]{Angelos Zavras}
\author[2]{Dimitrios Michail}
\author[3]{Christos Tryfonopoulos}
\affil[1]{Institute of Astronomy, Astrophysics, Space Applications \& Remote Sensing, National Observatory of Athens}
\affil[2]{Department of Informatics \& Telematics, Harokopio University of Athens}
\affil[3]{Department of Informatics \& Telecommunications,
University of the Peloponnese}


\maketitle

\begin{abstract}
The availability of the sheer volume of Copernicus Sentinel-2 imagery has created new opportunities for exploiting deep learning methods for land use land cover (LULC) image classification at large scales.
However, an extensive set of benchmark experiments is currently lacking, i.e. deep learning models tested on the same dataset, with a common and consistent set of metrics, and in the same hardware.
In this work, we use the BigEarthNet Sentinel-2 multispectral dataset to benchmark for the first time different state-of-the-art deep learning models for the multi-label, multi-class LULC image classification problem, contributing with an exhaustive zoo of 60 trained models.    
Our benchmark includes standard Convolution Neural Network architectures, as well as non-convolutional methods, such as Multi-Layer Perceptrons and Vision Transformers. We put to the test EfficientNets and Wide Residual Networks (WRN) architectures, and leverage classification accuracy, training time and inference rate. 
Furthermore, we propose to use the EfficientNet framework for the compound scaling of a lightweight WRN, by varying network depth, width, and input data resolution. Enhanced with an Efficient Channel Attention mechanism, our scaled lightweight model emerged as the new state-of-the-art. It achieves  4.5\% higher averaged F-Score classification accuracy for all 19 LULC classes compared to a standard ResNet50 baseline model, with an order of magnitude less trainable parameters. We provide access to all trained models, along with our code for distributed training on multiple GPU nodes. This model zoo of pre-trained encoders can be used for transfer learning and rapid prototyping in different remote sensing tasks that use Sentinel-2 data, instead of exploiting backbone models trained with data from a different domain, e.g., from ImageNet. We validate their suitability for transfer learning in different datasets of diverse volumes. Our top-performing WRN achieves state-of-the-art performance (71.1\% F-Score) on the SEN12MS dataset while being exposed to only a small fraction of the training dataset.
\end{abstract}

\section{Introduction}
\label{sec:intro}

\label{sec:introduction}
The Copernicus program is believed to be a game changer for Earth Observation (EO) science. Free and open data available at this scale, frequency, and quality constitute a fundamental paradigm change in EO \citep{nerc523287}. Today, Copernicus is producing 20 terabytes of satellite data every day, however, the availability of the sheer volume of Copernicus data outstrips our capacity to extract meaningful information.
Motivated by the success of deep learning (DL) methods in various data-intensive tasks e.g., in medicine~\citep{Esteva2019}, self-driving cars~\citep{Maqueda_2018_CVPR}, image classification~\citep{DBLP:journals/corr/abs-1712-04621}, machine translation~\citep{DBLP:journals/corr/abs-1803-07416}, etc., the remote sensing community has been exploiting deep learning methods to propel the research and development of new applications at scale~\citep{8113128}.

One prominent application for remote sensing (RS) imagery is land use / land cover (LULC) classification. Research has focused on both pixel-based~\citep{KHATAMI201689} and object-based~\citep{rs7010153} approaches. LULC mapping scale may vary from high-resolution~\citep{TONG2020111322} to global scale, e.g. the Copernicus Global Land Cover~\citep{rs12061044}. Machine learning~\citep{rs12071135}, and DL~\citep{7891032} methods, in particular, have been widely adopted by the community to classify LULC, with some studies exploiting the multi-temporal nature of satellite data, e.g.~\cite{8006221}. A particular problem family is multi-label LULC scene categorization~\citep{8633359}. The objective of image scene classification and retrieval is to automatically assign class labels to each RS image scene in an archive, and differs from semantic segmentation tasks for LULC mapping and classification. Adopting DL approaches for RS image scene classification problems have shown excellent performance~\citep{8730481}.

However, DL leads to highly nonlinear, generally overparameterized models~\citep{du2019gradient} that are prone to overfit. In order to use DL models that generalize well for previously unseen test data, we need to train them with large amounts of input labeled data. The data-hungry nature of modern machine learning has thus been a barrier for its widespread application in geosciences and RS. The lack of curated datasets and pretrained models tailored to RS data has prevented the use of traditional transfer learning approaches for LULC image classification. Therefore, researchers have used models pre-trained on optical datasets~\citep{sumbul2020bigearthnet}, such as ImageNet~\citep{deng2009imagenet}, to facilitate the training of new RS models using smaller labeled datasets. This kind of learned knowledge, however, cannot fully transfer on such a different data distribution. Features learned from optical datasets have different characteristics from the ones found in multispectral satellite imagery, therefore in principle, DL models should be trained from scratch to encode this information. 

In order to address the problem of the scarcity of labeled data for training DL models for LULC image classification,~\cite{8900532} created and published BigEarthNet, a large, labeled dataset, which contains single-date Sentinel-2 patches for multi-label, multi-class LULC scene classification. BigEarthNet is a benchmark dataset that consists of 590,326 Sentinel-2 image patches acquired between June 2017 and May 2018 over the 10 European countries, with spectral bands at 10, 20, and 60-meter resolution. Each image patch is annotated by the multiple land-cover classes (i.e., multi-labels) provided by the CORINE Land Cover (CLC) database of the year 2018~\citep{CLC2018} based on its detailed Level-3 class nomenclature. In order to increase the effectiveness of BigEarthNet, the authors introduced an alternative class-nomenclature to better describe the complex spatial and spectral information content of the Sentinel-2 imagery. The new classes nomenclature consists of 19 LULC classes~\citep{sumbul2020bigearthnet}. Recently, the dataset was enriched with Synthetic Aperture Radar Sentinel-1 patches~\citep{9552024}.

We identify three major gaps in LULC scene classification with DL, which we address in our work. 
First is gap is reproducibility, reusability and provenance of the trained models. Currently, an extensive set of benchmark experiments is lacking, i.e. DL models tested on the same dataset, with a common and consistent set of metrics, and in the same hardware. Published works on BigEarthNet (Section~\ref{sec:LULCmethods}) are fragmented, and in the absence of baseline studies, it is difficult to appreciate which methods work best.
Second is the testing and reporting of results on new, state-of-the-art, model architectures what have shown great promise in non-RS, Computer Vision applications. Given the rapid growth of DL research in this field, new families of approaches have emerged, other than traditional Convolutional Neural Networks (CNN), that is worth exploring in RS. 
Third is accounting for training efficiency and inference time, in addition to classification accuracy, as critical parameters that define overall model performance. This is especially important for training on large datasets, such as on BigEarthNet ($\sim$ 66Gb, $\sim$ 0.5 million image patches) or other similar training datasets for LULC classification, e.g.~\cite{hong2021multimodal, helber2019eurosat}. The sheer size of such datasets leads to significant training time overheads, which becomes a bottleneck for testing different model architectures and ideas. Therefore, new methods for efficient training are required to allow researching, engineering and fine-tuning novel DL architectures, including ablation studies and hyperparameter optimisation.


To address these gaps we rigorously benchmark DL models using the BigEarthNet dataset, analyzing their overall performance under the light of both speed (training time and inference rate) and model simplicity vis-à-vis LULC image classification accuracy. We investigate standard architectures, such as CNNs, and test novel, non-convolutional methods, such as Multi-Layer Perceptron (MLP) and Vision Transformer (ViT). To the best of our knowledge, it is the first time that ViT and MLP are used to encode multispectral information for LULC, setting-up a challenging state-of-the-art for future methods. ViTs in particular, are inherently data-hungry. Compared  to standard CNNs, the lack of inductive bias renders their training from scratch a way more difficult task, and are therefore difficult to be utilized in tasks with small datasets. These models are typically pretrained in large datasets and then finetuned for the task at hand \citep{DBLP:journals/corr/abs-2010-11929, steiner2021train}. Incorporating them in our model zoo, we finally make them available for exploitation in the remote sensing domain.

In addition, in order to address the requirement for efficiency in training, we explore lightweight architectures with very few parameters compared to typical CNNs. We focus on the use and adaptation of the framework for scaling EfficientNet~\citep{pmlr-v97-tan19a} encoders, and apply it to the order of magnitude more lightweight Wide Residual Network-WRN~\citep{DBLP:journals/corr/ZagoruykoK16}. Coupled with an implementation for efficient distributed training on 20 GPUs, we are able to experiment with several variations of such scalable models. Our benchmark identifies a set of novel, efficient, models, which are on par or better for most accuracy metrics with other published works, and with considerably fewer trainable parameters and memory requirements at both training and inference. The benchmark concludes with a new WRN model, enhanced with a spatio-spectral attention mechanism, which achieves the best overall performance and sets the new state-of-the-art (SOTA) for LULC image classification on the BigEarthNet. 


 



Our main contributions can be summarised as follows:
\begin{itemize}
    \item We benchmark 60 DL models for the task of multi-label, multi-class LULC single image classification. 
    
    \item We provide a DL model zoo based on Sentinel-2 data. The models and the implementation of our framework can be found on the project's github repository\footnote{https://github.com/Orion-AI-Lab/EfficientBigEarthNet}. We also provide the first pretrained Vision Transformer and MLP-mixer networks for multispectral Sentinel-2 data.
    
    \item We design and scale a new family of models based on Wide Residuals Networks (WRN) that follow the EfficientNet paradigm for scaling. This is the first time that WRN model compound scaling is applied in a remote sensing context and our results show great promise for performance enhancement in satellite image classification tasks. 
   
    \item We provide the new SOTA on the BigEarthNet dataset in terms of classification accuracy, training time, and inference rate. Our champion model outperforms a baseline ResNet50 model for all 19 LULC classes, achieving 4.5\% higher F-score, having an order of magnitude less trainable parameters.
    

    \item We show that convolution-free and lightweight architectures (e.g MLPMixer) can have comparable performance with their convolutional counterparts.
    
    
\end{itemize}

\section{Related work}
\subsection{LULC  scene classification with BigEarthNet}
\label{sec:LULCmethods}
Recent works have used BigEarthNet to experiment and test DL models for LULC scene classification. In~\cite{9096309}, a multi-attention strategy that utilizes a bidirectional long short-term memory network is adopted to capture and exploit the spectral and spatial information content of RS imagery. A new study by~\cite{9412588} proposes an oversampling method to cope with the BigEarthNet LULC class imbalance, while in~\cite{DBLP:journals/corr/abs-2105-05496} the authors propose a consensual collaborative multi-label learning method for harmonizing the BigEarthNet labels. In~\cite{kakogeorgiou2021evaluating} the authors use the DenseNet~\citep{Huang_2017_CVPR} model and test different Explainable Artificial Intelligence methods to interpret model predictions. Finally, in~\cite{9537619} a deep representation learning framework on fused BigEarthNet spectral bands is proposed for the same task. 

The BigEarthNet dataset has also been used for unsupervised and/or weekly supervised tasks. In ~\cite{sumbulinformative} a Deep Metric Learning framework is adopted, where a triplet sampling method is proposed to learn quality feature representations towards content-based image retrieval. In~\cite{rs12020207}, a U-Net~\citep{10.1007/978-3-319-24574-4_28} image classifier transferred to segmentation is trained with weak labels, outperforming pixel-level algorithms. The authors in~\cite{manas2021seasonal} show that pre-training with contrastive learning on BigEarthNet outperforms ImageNet pretrained models for LULC scene classification, while in~\cite{stojnic2021self} contrastive multiview coding is adopted for self-supervised pretraining. Similarly, in~\cite{9413112} colorization is proposed as a solid pretext task before using BigEarthNet labels for the LULC scene classification downstream task. 

\subsection{LULC scene classification with other datasets}
\label{sec:otherdata}

The work of~\cite{maggiori2016convolutional} has been one of the early works on LC classification with high-resolution RS images using transferable deep models. Since then, deep learning has been extensively used for LULC image classification, in different setups and for various datasets. SEN12MS by~\cite{Schmitt2019} is a dataset consisting of 180,662 triplets of dual-pol synthetic aperture radar (SAR) image patches, multispectral Sentinel-2 image patches, and MODIS land cover maps as labels. EuroSAT~\citep{helber2019eurosat} is another single label LULC image classification dataset, and is comprised of ten classes with a total of 27,000 labeled and geo-referenced  Sentinel-2 multispectral images. Finally, a typical  dataset used for deep learning based LULC high resolution RS scene classification is the well-known UC Merced (UCM) dataset by~\cite{yang2010bag}. The UCM dataset consists of 21 different labeled classes, but is relatively small in size, with only 100 images per class, which must then be divided between training and validation sets. 

To address the challenge of few training samples for DL, \cite{scott2017training} test a transfer learning with fine-tuning approach and a data augmentation strategy tailored specifically for remote sensing imagery for the UCM dataset. Finding efficient data augmentations is also the focus of~\cite{stivaktakis2019deep} for the same dataset. \cite{gomez2021msmatch} on the other hand, adopt a semisupervised learning approach to deal with label scarcity, which is tested on both the UCM and EuroSAT datasets. The approach is based on a combination of weak and strong data augmentations along with pseudolabeling. A semi-supervised processing chain based on the appropriate selection of labeled samples through a teacher model is proposed by~\cite{fan2020semi}, leveraging the availability of large amounts of unlabeled very high-resolution RS ShenzhenLC city data, for urban LULC image classification. The heterogeneous urban LC types of the city of Southampton, UK and its surrounding environment were used by~\cite{zhang2018hybrid}, proposing an MLP-CNN ensemble classifier for capturing deep spatial feature representations spectral discriminative information, for aerial imagery classification.

\cite{chaib2017deep} use traditional CNN models for feature extraction, while for the classification the authors rely on Support Vector Machines. This simpler approach is tested on the UCM dataset and the Aerial Image dataset~\citep{7907303}, a large-scale data set for aerial scene classification with more than 10,000 aerial scene images, annotated with 30 classes. \cite{TONG2020111322} setup the Gaofen Image Dataset (GID), a large-scale LC annotated dataset with Gaofen-2 (GF-2) satellite images. The authors exploit GID to pretrain standard CNNs models that capture the contextual information contained in different LC types, and propose a domain adaptation strategy by creating and appropriately selecting pseudolabels from a different target domain of unlabeled high resolution RS images. The transferability of their DL models is showcased on several datasets, including Gaofen-2, Gaofen-1, Jilin-1, Ziyuan-3, Sentinel-2A, and Google Earth platform data. High resolution RS data are also used by~\cite{lee2020different}, who propose a spectral domain transformation strategy on individual Landsat-8 multi-temporal pixel data, for creating two-dimensional matrices on which CNN models can be applied for LULC classification.  

Finally, \cite{zhang2020scale} address the issue of input image resolution, similarly to Efficient scaling, and develop an approach to automatically design a pyramid-like scale sequence that is fed to CNN models for aerial digital photography LULC image classification. Learning multiscale deep representations was also proposed by~\cite{zhao2016learning} for classifying RS images, an approach that was tested for three custom very high resolution RS datasets.


\subsection{EfficientNets in remote sensing}
EfficientNet is a family of DL models that are scaled to balance network depth, width, and input data resolution to achieve an optimal performance-training time trade-off. Before the EfficientNets came along, the most common way to scale up CNNs was either by one of three dimensions: 
\begin{itemize}
    \item Depth (number of hidden layers) as in~\cite{He_2016_CVPR}: although deeper networks tend to provide better image classification accuracy, they are also more difficult to train due to the well-known vanishing gradients problem. Accuracy gains quickly diminish beyond a certain depth. 
    \item Width (number of channels/filters) as in~\cite{DBLP:journals/corr/ZagoruykoK16}: while easier to train and able to capture fine-grained features, they encounter difficulties in capturing higher-level image content.
    \item Image resolution (image size) as in ~\cite{huang2019gpipe}: enhanced resolution of the input imagery in principle provides the CNN with more information. 
\end{itemize}
EfficientNets on the other hand perform Compound Scaling, i.e. scale simultaneously all three dimensions, depth, width, and image resolution, while maintaining a balance between all dimensions of the network. 

In RS, EfficientNets have lately gained traction, however in most cases they are used as a lightweight CNN backbone without care on how to scale them for the problem at hand. In~\cite{9320487} for example, the authors engineer a new CNN model that is based on the pretrained EfficientNet-B3 scaled CNN, enhanced with an attention mechanism for RS image classification. EfficientNet-B0 lightweight backbone with a recurrent attention module is employed for the same problem in~\cite{https://doi.org/10.1049/ipr2.12139}. EfficientNet-B0 and its deeper EfficientNet-B3 version are used in~\cite{rs11242908} for fine-tuning pre-trained CNNs, while in~~\cite{9129743} EfficientNet is used as a feature extractor coupled with a set of Softmax classifiers for knowledge adaptation across multiple RS sources. Semantic segmentation of high-resolution RS images is addressed with an EfficientNet-B1 model as lightweight network with attention modules in~\cite{9324723}. An EfficientNet with a reduced number of parameters but improved performance is proposed in~\cite{rs12101670} for mapping buildings damaged by a wide range of disasters. In~\cite{gomez2021msmatch}, the authors deploy EfficientNet models for semi-supervised multi-spectral scene classification with few labels. Finally, in~\cite{tian_resolution-aware_2020} EfficientNet-B0 is used as a backbone, mixed with an attention module, for RS object detection.

 Model compound scaling with EfficientNets have been sporadically adopted for addressing remote sensing applications. In~\cite{9454686}, the authors test and report performance for five different scaled versions of EfficientNet (B0-B4), to classify airport buildings within Google Earth collected and annotated RGB imagery. \cite{WU2020106132} also use Google Earth imagery for an aircraft type recognition task, and apply compound scaling for balancing network width, depth, and resolution and selecting the best performing EfficientNet. In~\cite{isprs-archives-XLIII-B2-2020-731-2020} the authors use EffcientNet as a backbone feature encoder to their network for delineating buildings and argue that employing the compound scaling method allows the scaled model to focus on more relevant regions with enhanced object details. Similarly, \cite{rs14010106} scale an Efficient based model for pavement defect detection and classification. They show preference for the EfficientNetB4 model which although has marginally lower detection accuracy with respect to the EfficientNetB5 counterpart, the former has fewer trainable parameters. Finally, \cite{rs14030516} use and scale EfficientDet~\cite{tan2020efficientdet} that combines EfficientNet architecture with a bi-directional feature pyramid network for object detection in very high resolution satellite imagery.

\subsection{Wide Residual Networks in remote sensing}
 WRNs have been used to a lesser extent than EfficientNets to address remote sensing applications. They have been mainly used as part of benchmark studies and compared with traditional CNNs, e.g. as by~\cite{https://doi.org/10.48550/arxiv.2001.09614} for remote sensing image classification. Similarly,~\cite{s21238083} show that a WRN encoder has superior performance than other CNNs for LULC classification using the EuroSAT dataset~\citep{helber2019eurosat}. A variant of WRN was also used by~\cite{9173783} as an encoder to construct feature embeddings that are then passed onto the nodes of a graph neural network for multi-label remote sensing image classification. 

More complex approaches that build on WRNs have also been researched. \cite{8344565} test different 3-D architectures for airborne image classification considering the WRN trade-offs between network width versus depth. \cite{8126255} adopt a modified WRN, with wider convolutional channels and fewer network layers, to identify tsunami induced damages in build-up areas from Synthetic Aperture Radar data. \cite{DIAKOGIANNIS202094} focus on a semantic segmentation application, split into sequential tasks and addressed with an hierarchical model. The authors perform an ablation study that follows the WRN philosophy for understanding the performance gains, considering both model complexity and training convergence. \cite{8917718} build on the WRN concept to propose a new residual unit to limit diminishing feature reuse, as indicated by~\cite{DBLP:journals/corr/ZagoruykoK16}. Finally, \cite{8898850} work on hyperspectral image classification and propose an attention transfer architecture for domain adaptation between two WRNs.

\subsection{Attention modules in remote sensing}

Deep learning architectures that incorporate attention modules have been widely used in remote sensing and have shown to provide improvements for different applications, e.g.  for image classification, image segmentation, change detection, and object detection. The main variations include spatial, temporal, channel, cross and self-attention networks, while an overview of the main attention mechanism approaches used in RS is provided by~\cite{ghaffarian2021effect}.

To cope with the large receptive fields of traditional CNNs, ~\cite{ding2020lanet} propose a complex attention structure, that focuses on both low level features extracted from the early layers of a CNN, and high level features extracted from the late layers. The architecture is able to enrich the semantic information of low-level features by embedding local focus from high-level features and the algorithm is applied for RS image segmentation. Similarly,~\cite{9324913} propose a spatial and channel attention consistent model to capture both local and global features for RS image classification. For the same task,~\cite{8454883} design a recurrent attention mechanism that is able to fit high-level semantic and spatial features into simple representations, managing to accelerate the convergence rate and improve the classification accuracy. \cite{alhichri2021classification} enhance the EfficientNet-B3 encoder with a variation of the Squeeze-and-Excitation attention mechanism~\citep{Hu_2018_CVPR}, that we also test in our work, for RS image classification. Squeeze-and-Excitation channel attention mechanism is also applied by~\cite{tong2020channel} for the same problem, using a different encoder. \cite{zhao2020remote} propose two simple spatial and channel attention modules, which are able to reduce the impact of many small objects and complex backgrounds on RS image classification. 

\cite{8513990} focus on object detection for VHR RS imagery, and develop an architecture comprising of an encoder-decoder model that extracts features at multiple scales, followed by a different, trainable, attention network for each scale. A multi-scale approach is also adopted by~\cite{rs12101662} for RS image change detection, in which multiple spatial-temporal attention networks are trained to capture such dependencies at various scales. 

Self-attention approaches have gained traction in RS. \cite{cao2020self} propose a spatial and channel nonparametric self-attention layer, to enhance the semantic information propagated from representative objects, for RS image classification. A self-attention model is designed by~\cite{wu2020self} to reduce the interference of complex backgrounds and to focus on the most salient region of each image, also for RS image classification. \cite{martini2021domain} exploit Sentinel-2 time-series and develop a self-attention-based network tailored for domain adaptation for LC and crop classification in different geographic regions. Finally, in contrast to self-attention approaches where the input is a single embedding sequence, cross-attention combines asymmetrically two separate embedding sequences of the same dimension, e.g. as in~\cite{cai2020remote} where the authors develop a cross-attention mechanism for hyperspectral data classification, showing improved performance on several popular relevant RS data sets.

\section{Models in the benchmark}
\begin{figure*}[ht]
 \includegraphics[width=2\columnwidth]{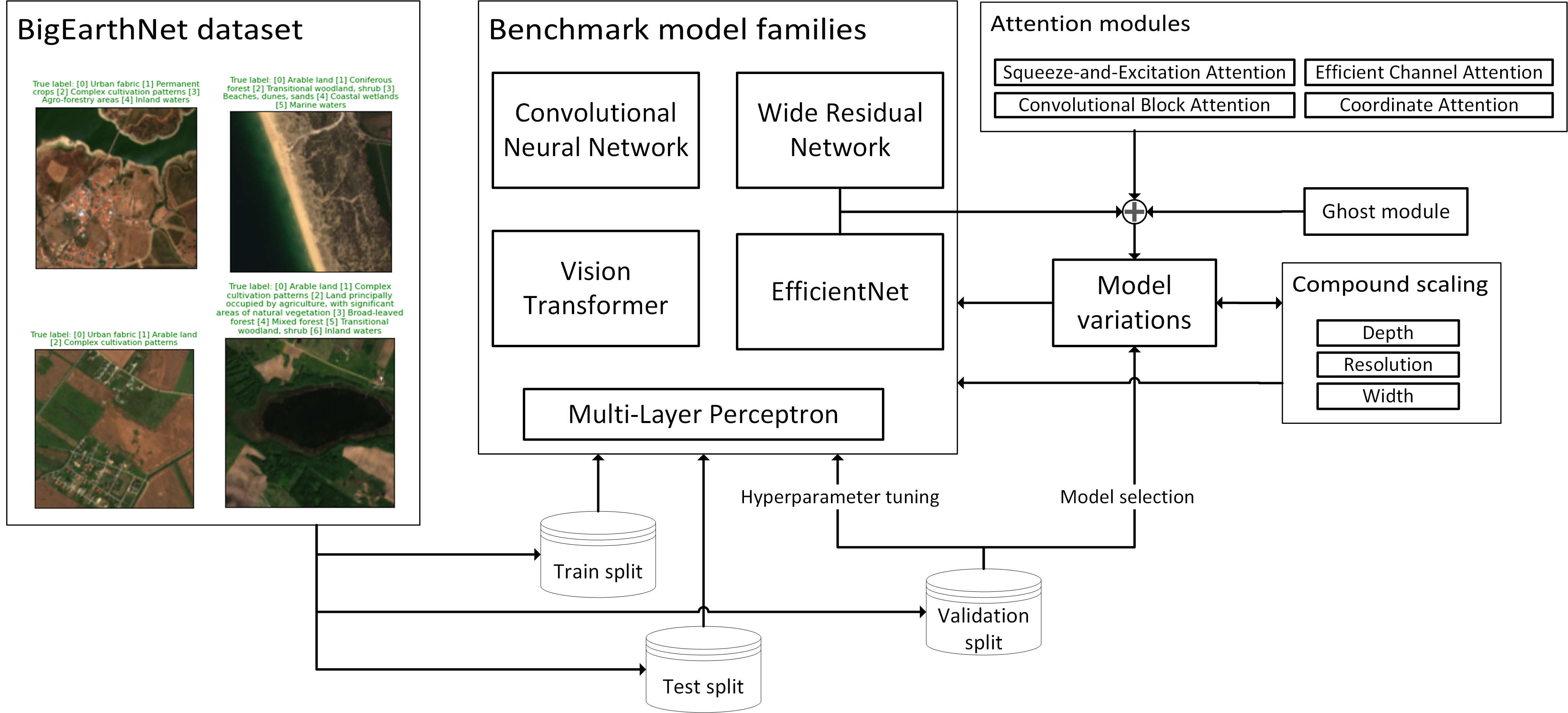} 
\caption{ Workflow for the benchmark. There are five model families tested in the benchmark. For EfficientNets and Wide Residual Networks we test variations with different attention modules and the addition of a Ghost module. We select the best performing model from each of the two base architectures and scale them through compound scaling. EfficientNet and Wide Residual Network architectures are explained in Figure~\ref{fig:EfficientNetArch} and Figure~\ref{fig:WideResNetArch} respectively.}
\label{fig:methodology}
\end{figure*}

In this section, we present the DL models that we deploy and benchmark for LULC scene classification.
These models are tested by adapting and customizing state-of-the-art DL architectures in Computer Vision, which have not yet been evaluated in remote sensing for the particular task.  Figure~\ref{fig:methodology} summarises the workflow of the benchmark.

\subsection{Vision transformer}
\label{sec:vit}
Transformers~\citep{DBLP:journals/corr/VaswaniSPUJGKP17} are a typical example of an architecture that makes the most of attention mechanism. These architectures have been successfully deployed in tasks concerning natural language processing~\citep{devlin2018bert}. This success has driven the research community to extend the traditional transformer architecture to computer vision. Recently, transformers have been successfully tested for hyperspectral image classification~\citep{9565208}. 
The Vision Transformer (ViT) in~\cite{DBLP:journals/corr/abs-2010-11929} processes images as follows. First, it splits the input image in N non-overlapping patches, and each patch (token) is linearly embedded. A class embedding is prepended to the token sequence, while positional embeddings are added to the patch embeddings to ensure positional information is not lost. 

In our implementation, we use a fully connected layer for the encodings. The output of this process is the input to a standard Transformer encoder. The classification is based on the prepended class token or a global average pooling of all tokens if the class token was not prepended~\citep{arnab2021vivit}. 
 We examine 5 versions of the ViT architecture. The first four are identical, and we simply vary the patch size. These models will be referred to as ViT/PatchSize e.g ViT/20. They consist of 8 transformer layers with 4 attention heads. Additionally, we examine a ViT with 12 transformer layers, 10 attention heads, and a patch size equal to 20. It will be denoted as ViTM/20.

\subsection{Multi-layer Perceptrons}
\label{sec:mlp}
Ever since the attention mechanism gained popularity, multiple networks based on attention have emerged. Alternatives to this strategy have made their appearance though, with the reemergence of simple MLP architectures as efficient lightweight models. Recent work in~\cite{tolstikhin2021mlpmixer} has shown that a plain architecture based on MLPs can compete with complex CNN and Transformer architectures. 
The MLP-Mixer~\citep{tolstikhin2021mlpmixer} architecture, for example, splits the input in K patches (tokens) and produces an embedding for each one of them. The embeddings are then fed in $\times N$ Mixer Layers. The final prediction is produced by a simple classification head. Each Mixer Layer consists of two blocks: the token-mixing MLP block and the channel-mixing MLP block. 

In this work, we build on top of the MLP-Mixer for the fast training of lightweight models with high throughput. We use two versions of the MLP-Mixer in particular. The base version, which we call MLP-Mixer, uses patch size of 12, a hidden dimension of 128 for the linear embeddings, and  4 Mixer layers. Each layer uses channel MLP dimension of 200 and token MLP dimension of  64. The second version called MLPMixerTiny uses a patch size of 6, embedding hidden dimension of 30, and 2 Mixer Layers. The token MLP dimension is set at 12 and the channel MLP dimension at 50.
 Following \ref{sec:vit}, any MLPMixer variant with different patch size will be referred as MLPMixer/PatchSize.

\subsection{EfficientNet and WRN-based models}
\label{sec:variations}

We deploy in our benchmark the original model in~\citet{pmlr-v97-tan19a}, which introduces a new baseline CNN architecture called EfficientNet-B0. Based  on  MnasNet~\citep{Tan_2019_CVPR}, this baseline network uses the Inverted Residual Block (MBConv Block), as in~\cite{howard2017mobilenets}, a type of residual block used by several mobile-optimized CNNs for efficiency reasons, with the addition of a Squeeze-and-Excitation (SE) block~\citep{Hu_2018_CVPR}.

\begin{figure*}[ht]
 \includegraphics[width=2\columnwidth]{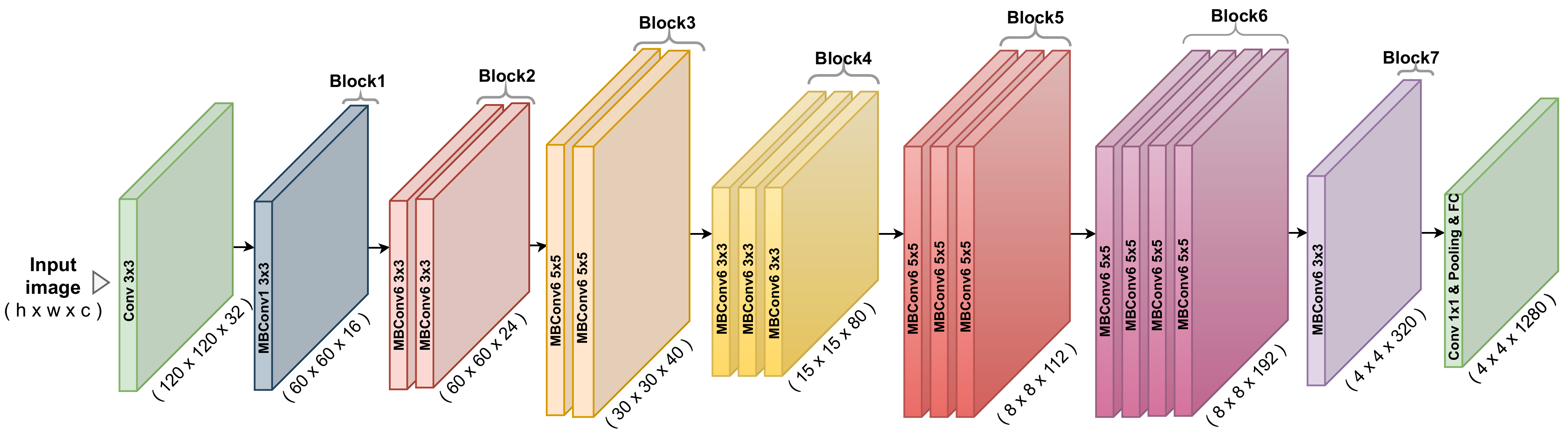}
\caption{Our EfficientNet base model architecture implemented in the benchmark. MBConv1 and MBConv6 blocks are explained in Figure~\ref{fig:EfficientNetModules}.}
\label{fig:EfficientNetArch}
\end{figure*}

\begin{figure}[h]
 \includegraphics[width=\columnwidth]{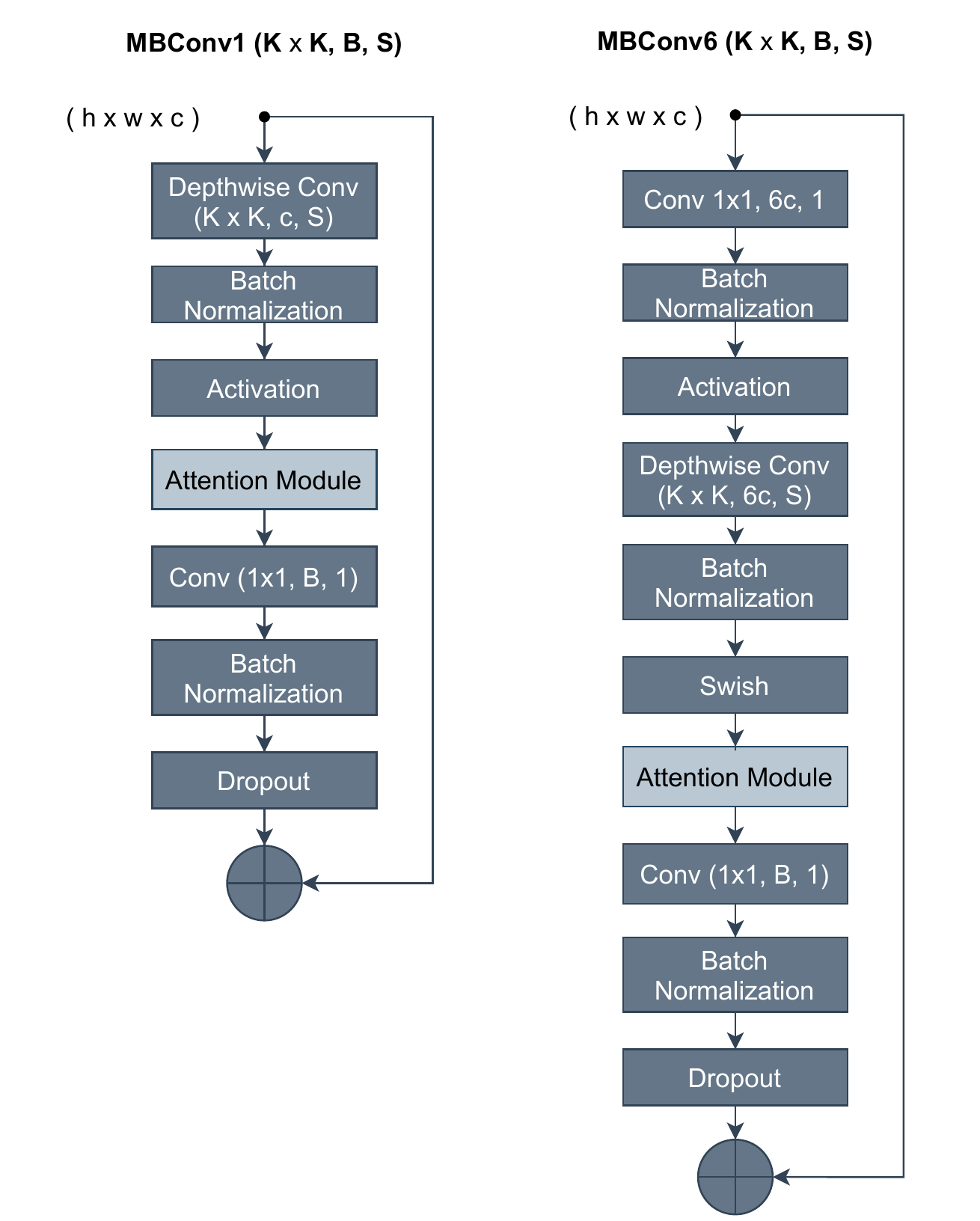}
 \centering
\caption{The modules of MBConv1 and MBConv6 blocks used in the architecture of Figure~\ref{fig:EfficientNetArch}. With light blue, we mark the position of the Attention mechanism, for which we substitute the different attention modules as explained in Section~\ref{sec:variations}.}
\label{fig:EfficientNetModules}
\end{figure}


Furthermore, we investigate the impact of yet another efficient CNN encoder, the WRN~\citep{DBLP:journals/corr/ZagoruykoK16}. Wide Residual Networks constitute an enhancement to the original Deep Residual Networks. Instead of relying on  increasing the depth of a Residual Network to improve its accuracy, it was demonstrated that a network could be made shallower and broader without compromising its performance. Prior to the introduction of WRNs, Deep residual networks (e.g., ResNets) were shown to have a fractional boost in performance, but at the cost of roughly doubling the number of layers. This led to the problem of diminishing feature reuse~\citep{DBLP:journals/corr/SrivastavaGS15} and overall made the models slower to train. WRNs showed on popular benchmark datasets that having a wider residual network, by widening the ResNet blocks, leads to better  performance  with  respect to deeper counterparts. In~\cite{DBLP:journals/corr/ZagoruykoK16}, the authors also show that  gradually increasing both depth and width helps until the number of parameters becomes too high and stronger regularization is needed.

\begin{figure*}[ht]
 \includegraphics[width=1.5\columnwidth]{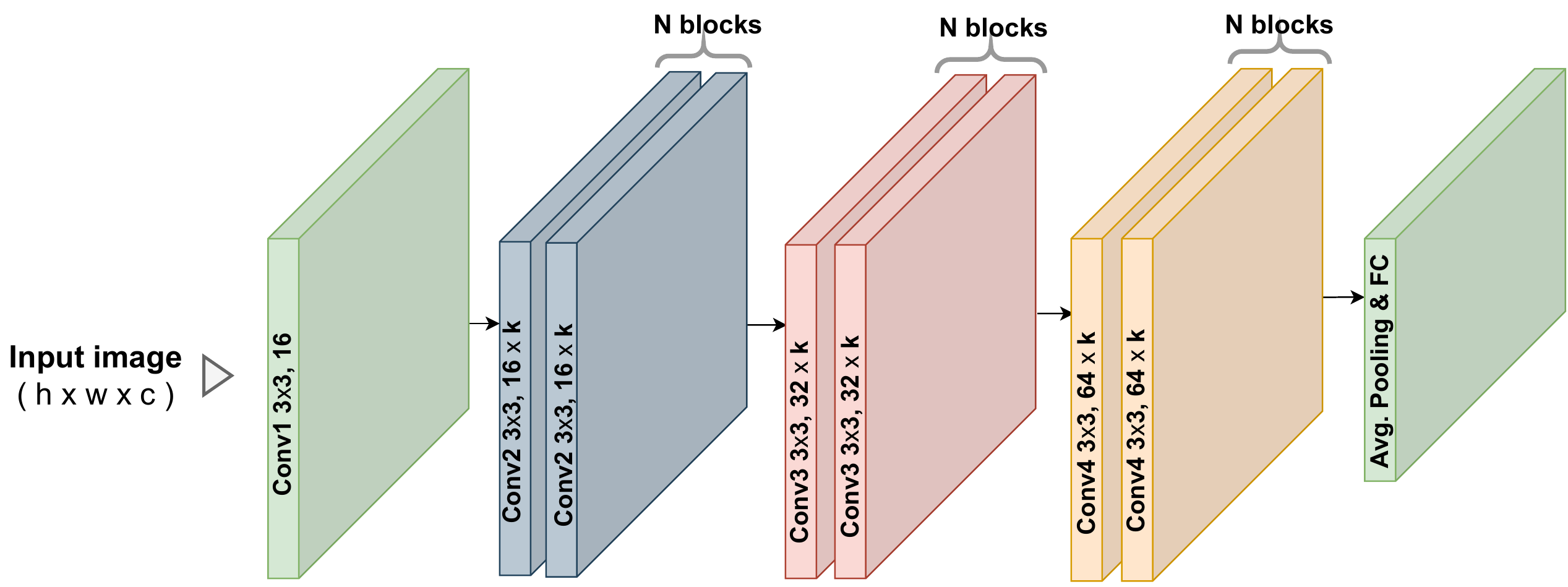}
 \centering
\caption{Our WRN base model architecture implemented in the benchmark. The blocks of the architecture are explained in Figure~\ref{fig:WRNModules}.}
\label{fig:WideResNetArch}
\end{figure*}

\begin{figure}[h]
 \includegraphics[width=\columnwidth]{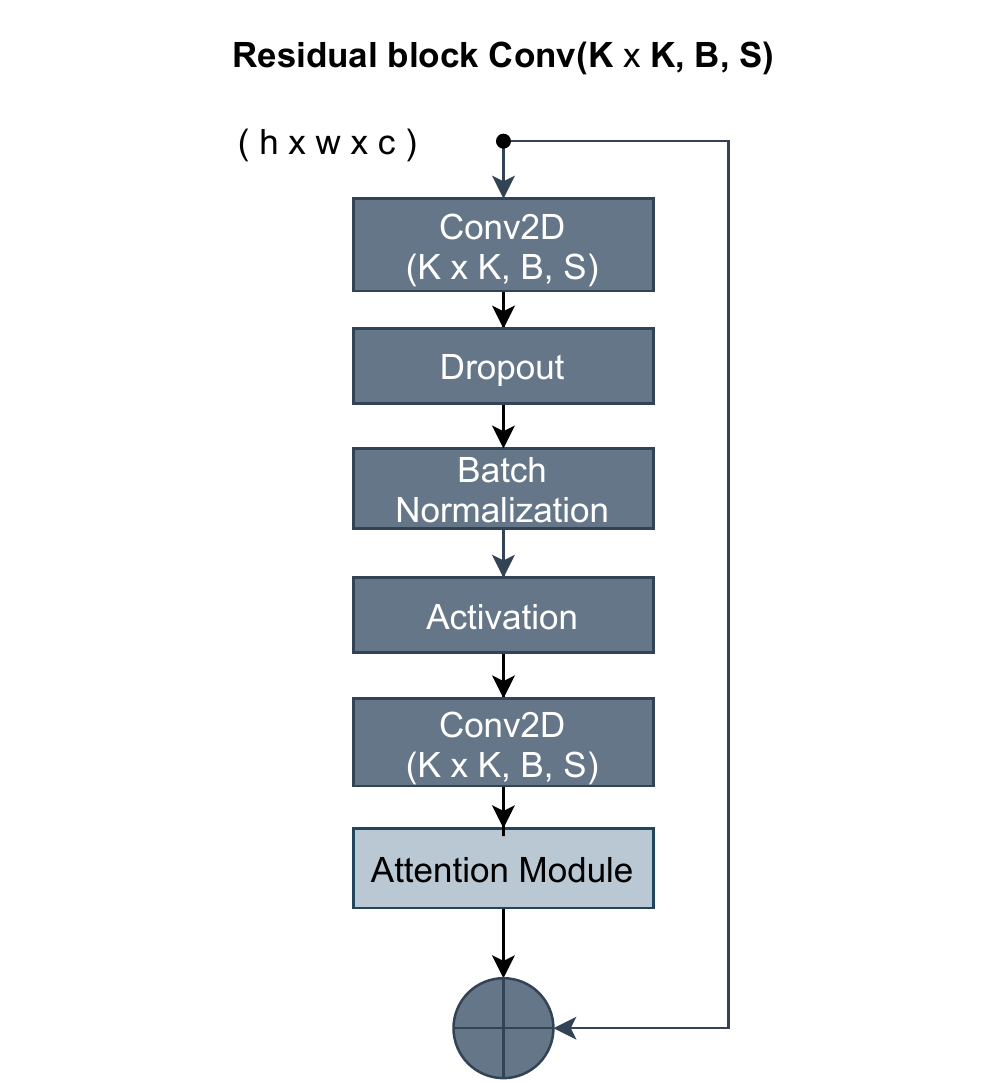}
 \centering
\caption{The modules of the residual block used in the architecture of Figure~\ref{fig:WideResNetArch}. With light blue, we mark the position of the Attention mechanism, for which we substitute the different attention modules as explained in Section~\ref{sec:variations}.}
\label{fig:WRNModules}
\end{figure}


We design and deploy two different base models in this benchmark. The first one uses EfficientNet-B0 as a backbone, enhanced with different attention mechanisms and a ghost module that we introduce in this section. The model architecture is presented in Figure~\ref{fig:EfficientNetArch}. The second base model uses WRN as a backbone and is also tested with different attention mechanisms and with the addition of a ghost module. As our baseline model we use the smallest WRN possible, WRN-10-2, which denotes a residual network that has a depth of 10 convolutional layers and a widening factor of 2. The architecture of the WRN-based family of models is presented in Figure~\ref{fig:WideResNetArch}. 

\subsubsection{EfficientNet/WRN with ghost module}

Inspired by the work in~\cite{Han_2020_CVPR}, we test the effect of a Ghost module for our base architectures. This refers to an essentially standalone replacement layer for standard convolution layers in deep neural network architectures. In principle, this alternative convolutional layer runs linear transformations on fewer feature maps and still fully reveals information of the underlying intrinsic features. The main functionality of the Ghost module we use is to remove redundant copies of unique intrinsic feature maps (Ghost Feature Maps) learned by different convolutional layers in deep networks, so that we preserve the feature-rich representations of the input image while avoiding redundant convolution operations. In this benchmark, we use the suffix $-$ghost to denote a model that uses a Ghost module in its architecture.




\subsubsection{EfficientNet/WRN with attention modules}
While conventional CNNs extract features by fusing spatial and channel information within local receptive fields, attention mechanisms can enhance the important parts of the input data either or both in the spatial and the spectral domain.
In our benchmark, we experiment with different spatial and channel attention mechanisms for both EfficientNet and WRN based models. The exact position of the attention mechanisms is shown in Figures~\ref{fig:EfficientNetModules} and~\ref{fig:WRNModules} for the two base architectures of Figures~\ref{fig:EfficientNetArch} and~\ref{fig:WideResNetArch} respectively.


Firstly, we evaluate the Squeeze-and-Excitation Attention Module (SE), as in~\cite{Hu_2018_CVPR}, a channel attention building block, facilitating dynamic channel-wise feature recalibration via channel-wise dependency modeling. Given the input features, the SE block first employs a 2D Global Average Pooling (GAP) for each channel independently, then two fully-connected layers with non-linearity followed by a sigmoid function are used to generate channel weights. The two fully-connected layers are designed to capture non-linear cross-channel interactions, which involves dimensionality reduction for controlling model complexity. Hereinafter, we use the suffix \textit{$-$SE} to denote a model that uses a Squeeze-and-Excitation Attention module in its architecture.

Channel attention mechanisms, such as the SE, have demonstrated promising results in the design of lightweight mobile networks. Nevertheless, a fundamental shortcoming of those mechanisms is that positional attention information is neglected, which is critical for vision tasks. Later works, such as the Convolutional Block Attention Module (CBAM)~\cite{Woo_2018_ECCV}, attempt to exploit positional information with little to no additional computational cost by reducing the channel dimension of the input tensor and then computing spatial attention using convolutions. CBAM consists of two consecutive sub-modules, the Channel Attention Module (CAM) and the Spatial Attention Module (SAM). CAM is similar to the SE with a small modification. Instead of reducing the Feature Maps to a single pixel by GAP, it decomposes the input tensor into 2 subsequent vectors of dimensionality (c $\times$ 1 $\times$ 1). One of these vectors is generated by GAP while the other vector is generated by Global Max Pooling (GMP). Average pooling is mainly used for aggregating spatial information, whereas max pooling preserves much richer contextual information in the form of edges of the object within the image, which leads to finer channel attention. The authors validate this in their experiments where they show that using both GAP and GMP gives better results than using just GAP as in the case of SE. SAM, on the other hand, is a three-step sequential procedure. The first phase is called the Channel Pool and it contains max pooling and average pooling operations applied across the channels to the input (c $\times$ h $\times$ w), to generate an output with shape (2 $\times$ h $\times$ w). This is the input to a convolution layer that outputs a 1-channel feature map (1 $\times$ h $\times$ w). After feeding the output into a BatchNorm and an optional ReLU, the data enter a Sigmoid Activation layer. Finally, the attention maps produced by CAM and SAM are multiplied with the input feature map for adaptive feature refinement. In this work, we use the suffix \textit{$-$CBAM} to denote a model that uses a Convolutional Block Attention Module in its architecture.

CBAM adopts convolutions to capture local relations but fails to model long-range dependencies. In order to deal with this drawback,~\citep{Hou_2021_CVPR} proposed Coordinate Attention (CA) module, a novel attention mechanism which embeds positional information into channel attention so that the network can focus on large important regions at little computational cost. CA captures long-range spatial dependencies while alleviating positional information loss, caused by the 2D global pooling, by factorizing channel attention into two parallel 1D feature encoding processes that effectively integrate spatial coordinate information into the generated attention maps. More precisely, the CA approach uses two 1D global pooling operations to aggregate the input features in the vertical and horizontal directions into two separate direction-aware feature maps. These two feature maps with embedded direction-specific information are then independently encoded into two attention maps, each of which captures long-range dependencies of the input feature map along one spatial direction. As a result, the positional information can be preserved in the generated attention maps. Both attention maps are then applied to the input feature map via multiplication to emphasize the representations of interest. We use the suffix \textit{$-$CA} to denote a model that uses a coordinate attention module in its architecture.

Although the strategy of dimensionality reduction for controlling model complexity is widely used in the aforementioned attention modules,~\cite{wang2020ecanet} claim that dimensionality reduction has side effects on channel attention prediction and it is inefficient and unnecessary to capture dependencies across all channels. Therefore, they propose the Efficient Channel Attention (ECA) module, which avoids dimensionality reduction and captures cross-channel interaction in an efficient way, so that both efficiency and effectiveness are preserved. It first performs channel-wise global average pooling and then captures channel attention through a fast $1D$ convolution, whose kernel size is adaptively determined by a non-linear mapping of the channel dimension. The ECA block models local cross-channel interaction by considering every channel and its $k$ neighbors. We use the suffix \textit{$-$ECA} to denote a model that uses an Efficient Channel Attention module in its architecture.

\subsection{Method for scaling-up our EfficientNet and WRN models designs}
\label{sec:scaleup}

Our framework for scaling the base models of Figures~\ref{fig:EfficientNetArch} and~\ref{fig:WideResNetArch}, and their variants with the different attention mechanisms and the ghost module, is the compound model scaling method, as in~\cite{pmlr-v97-tan19a}. It consists of a set of rules to scale the three dimensions of our model architectures, depth, width, and input data resolution, using a compound coefficient $\phi$. Through that coefficient, dimensions do scale uniformly. $\phi$ represents how many more resources are available for the model to scale, while $\alpha$, $\beta$, $\gamma$ are parameters that assign those extra resources to the dimensions of the network: 

\begin{equation}
    \label{eq:compound}
    \begin{aligned}[b]
    & depth \;\; d = a^{\phi}\\
    & width \;\; w = \beta^{\phi}\\
    & resolution \;\; r= \gamma^{\phi}\\
    & s.t. \;\; \alpha \cdot \beta^{2} \cdot \gamma^{2} \approx 2 \\
    & \alpha \geq 1,  \beta \geq 1, \gamma \geq 1,
    \end{aligned}
\end{equation}


 We start from a baseline EfficientNetB0 (Figure~\ref{fig:EfficientNetArch}) and WRNB0 (Figure~\ref{fig:WideResNetArch}) models and scale them in two steps: first we determine the $\alpha, \beta, and \gamma$ coefficients for our EfficientNet and WRN base models. For EfficientNet we use the same coefficients determined by grid-search and used by the authors. For WRN on the other hand, we perform a grid search to determine $\alpha, \beta,and \gamma$. At the same time, inspired by~\cite{bello2021revisiting}, we evaluate the effect of width scaling prioritization over depth, in contrast to the EfficientNet paradigm. It should be noted that due to the rounding of the number of filters and the number of blocks, some coefficients result in the same network. In such cases, we kept the coefficients resulting in the lowest $\alpha \cdot \beta^{2} \cdot \gamma^{2}$ product, based on Eq.~\ref{eq:compound}. We report the results of our grid search for WRN in Table~\ref{tab:WRNScalingCoeff} and summarize the coefficients used for scaling our EfficientNet and WRN base models in Table~\ref{tab:CompScalingCoeff}.
Second, we fix the aforementioned $\alpha, \beta, and \gamma$ and scale our baseline model by varying $\phi$, using Eq.~\ref{eq:compound}. Our baseline B0 models make use of 60$\times$60 resolution images, reaching up to 120$\times$120 for our B7 models. Eight models can be generated, from EfficientNetB0 up to EfficientNetB7. Similarly, another eight WRN models can be generated, from B0 to B7. To the best of our knowledge, we are the first to use WRNs, enhance them with different attention modules, and scale them using the EfficientNet paradigm.

\begin{table}[b]
    \centering
    \begin{tabular}{ c c c c c c}
    \toprule
        \multirow{1}{*}{{$\alpha$}} & $\beta$ & $\gamma$ & F-Score & Training Time  & Model Size \\
        & & & (\%) & (hours.mins) & \\
        \hline
        1.1 & 1.2 & 1.1 & \textbf{75.6} & 0.20 & 433,195\\
        \hline
        1.2 & 1.1 & 1.1 & 75.2 & 0.21 & 373,367\\
        \hline
        1.2 & 1.3 & 1.1 & 75.5 & 0.22 & 520,091\\
        \hline
        1.3 & 1.1 & 1.1 & 75.1 & 0.23 & 410,471\\
        \hline
        1.4 & 1.1 & 1.1 & 74.8 & 0.24 & 447,114\\
        \hline
        1.1 & 1.2 & 1.2 & 74.6 & 0.22 & 433,195\\
        \hline
        1.2 & 1.1 & 1.2 & 74.2 & 0.23 & 373,367\\
        \hline
        1.1 & 1.2 & 1.3 & 74.4 & 0.23 & 433,195\\
        \hline
        1.2 & 1.1 & 1.3 & 74.0 & 0.23 & 373,367\\
        \hline
        \bottomrule\\
    \end{tabular}
    \caption{ Grid-search used to determine the optimum compound scaling coefficients $\alpha, \beta, \gamma$, which are later used for scaling our WRN model.} 
    \label{tab:WRNScalingCoeff}
\end{table}

\begin{table}[b]
    \centering
    \begin{tabular}{ c c c c}
    \toprule
        Model & $\alpha$ & $\beta$ & $\gamma$ \\
        \hline
        EfficientNet & 1.2 & 1.1 & 1.1 \\
        \hline
        WRN & 1.1 & 1.2 & 1.1\\
        \hline
        \bottomrule\\
    \end{tabular}
    \caption{Compound scaling coefficients $\alpha, \beta, \gamma$ determined by grid-search and used for scaling our EfficientNet and WRN models.} 
    \label{tab:CompScalingCoeff}
\end{table}


\subsection{K-Branch CNN}
K-Branch CNN, is a variant of the attention based model introduced by~\cite{sumbul2019novel}. Assuming that different bands come with different spatial resolution, K-Branch CNN processes the bands grouped by spatial resolution in different branches. Each branch produces a local descriptor for the respective spatial resolution. To classify the input image, the respective descriptors are concatenated and fed to a fully connected layer. We have adapted the implementation provided by the authors to our framework to conduct our experiments.

\subsection{Traditional convolutional neural networks}
In our study, we include three traditional convolutional neural network families to serve as baselines i.e VGG, ResNet and DenseNet. These architectures have been widely used in both remote sensing e.g. \citep{sumbul2021bigearthnetmm,helber2019eurosat,kakogeorgiou2021evaluating} and computer vision applications. We briefly discuss the core ideas in the following subsections.
\subsubsection{VGG}
VGG~\citep{simonyan2015deep} is one of the reference convolutional neural networks. It received the first and second place in the ImageNet challenge 2014 in the localization and localization tracks respectively. It is designed to use small 3x3 convolutional filters, while increasing the depth of the neural network.

\subsubsection{Residual Neural Network}
Residual Neural Networks~\citep{He_2016_CVPR} have been the golden standard in multiple tasks for a long time. They introduced skip-connection blocks that enabled the training of very deep neural networks, up to 8x deeper than VGG and showed that increased depth can lead to considerable boost in accuracy, earning the first place in the classification track of the ImageNet challenge 2015.
\subsubsection{Dense Convolutional Neural Network}
In a similar fashion as ResNets, DenseNet~\citep{Huang_2017_CVPR} proposes to connect all layers directly instead of single skip connections. To achieve that, each layer receives as inputs the concatenated features of all preceding layers. A composition of such densely connected blocks with transition layers that perform downsampling (including pooling and convolutions) results to the Dense Convolutional Neural Network.

\section{Experiments}
\subsection{Dataset and experimental setup}
We use the BigEarthNet pre-defined splits as in~\cite{9552024} to train, test, and validate our models, based on the 19 land use classes nomenclature. This corresponds to 295,118 (33 GB), 147,559 (17 GB) and 147,559  (17 GB) Sentinel-2 image patches respectively. In addition to the 10$^{th}$ spectral band of Sentinel-2, which was also excluded in the original BigEarthNet implementation, we also excluded the 1$^{st}$ and 9$^{th}$ spectral bands as they do not contain information regarding the Earth’s surface.

We benchmark five model families in this work. First, we test traditional CNN models, including ResNets, DenseNets, and VGG. We then proceed to test the non-convolutional MLP models (Section~\ref{sec:mlp}). We use two versions of the MLPMixer~\citep{tolstikhin2021mlpmixer}, which we refer to as ``MLPMixer'' and ``MLPMixerTiny'', with 446,723 and 40,863 trainable parameters respectively. Transformer networks ViT~\citep{DBLP:journals/corr/abs-2010-11929}
are also trained and reported in our models zoo (Section~\ref{sec:vit}). We then benchmark EfficientNetB0 baseline network~\citep{pmlr-v97-tan19a} and test the impact of the attention mechanisms (squeeze and excitation, efficient channel attention, coordinate attention and convolutional block attention module) and variations with and without a ghost module, as described in Section~\ref{sec:variations}. A total of eight models is produced. Finally, we repeat the same set of experiments for WRNB0 baseline network and its variations. A total of ten models is produced, two more compared to EfficientNetB0, since the latter already contains the squeeze and excitation attention module in its baseline architecture. Based on these experiments, we select the best performing EfficientNetB0 and WRNB0 variation, considering both classification accuracy and training time metrics. These models are subsequently scaled-up from B0 up to B7 using the compound scaling methodology described in Section~\ref{sec:scaleup}. 

Given the computing resources available at the HPC infrastructure and depending on the model size, batch size varies between 32 and 256 and the learning rate varies between $0.00001$ and $0.001$ with a step decay at epochs 24 or 27. We train all models for a total of 30 epochs using the Adam optimizer. The learning rate is scaled by the number of workers in each run. The weights are initialized randomly. We select to minimize the Binary Cross Entropy loss function.

\subsection{Evaluation metrics}
\label{sec:metrics}
In supervised learning, for a multi-class problem, different methods are used to evaluate the generalization performance of a model, such as Accuracy and Area Under the Receiver Operating Characteristic (ROC) curve. In a multi-label setting, which is a generalization of multi-class classification, evaluating performance is more complicated than single-label classification problems, due to the simultaneous presence of multiple labels in the scene. Several problem transformation methods exist for multi-label classification. We adopt the binary relevance method~\citep{2011}, which entails training distinct binary classifiers, one for each label.
Each node in the output layer of our networks uses the sigmoid activation, so that a probability of class membership for the label is predicted. The results of each test sample can be assigned to one of the four categories:

\begin{itemize}
    \item True Positive (TP) - the label is positive and the prediction is also positive
    \item True Negative (TN) - the label is negative and the prediction is also negative
    \item False Positive (FP) - the label is negative but the prediction is positive
    \item False Negative (FN) - the label is positive but the prediction is negative
\end{itemize}

Here we define a set D of N examples and $Y_{i}$ to be a family of ground truth label sets and $P_{i}=h\left(x_{i}\right)$ to be a family of predicted label set. Following the formulation in~\cite{6471714}, the union set of all unique labels is:

\begin{equation}
L=\bigcup_{i=0}^{N-1} L_{i}
\end{equation}

While the definition of indicator function $I_{A}$ on a set A is presented as:

\begin{equation}
I_{A}(x)= \begin{cases}1 & \text { if } x \in A \\ 0 & \text { otherwise }\end{cases}
\end{equation}




Micro Precision (precision averaged over all label pairs) is defined as:
\begin{equation}
\label{eq:MicroP}
Pr_{micro}=\frac{T P}{T P+F P}=\frac{\sum_{i=0}^{N-1}\left|P_{i} \cap L_{i}\right|}{\sum_{i=0}^{N-1}\left|P_{i} \cap L_{i}\right|+\sum_{i=0}^{N-1}\left|P_{i}-L_{i}\right|}
\end{equation}

Micro Recall (recall averaged over all the label pairs) is defined as:
\begin{equation}
 \label{eq:MicroR}
R_{micro}=\frac{T P}{T P+F N}=\frac{\sum_{i=0}^{N-1}\left|P_{i} \cap L_{i}\right|}{\sum_{i=0}^{N-1}\left|P_{i} \cap L_{i}\right|+\sum_{i=0}^{N-1}\left|L_{i}-P_{i}\right|}
\end{equation}

Micro F Measure by label is the harmonic mean between Micro Precision and Micro Recall.
\begin{equation}
    \label{eq:MicroF1}
    \begin{aligned}[b]
    &\hspace{0.9cm}F_{micro} =  2\cdot \frac{T P}{2 \cdot T P+F P+F N}= \\=2\cdot
    & \frac{\sum_{i=0}^{N-1}\left|P_{i} \cap L_{i}\right|}{2 \cdot \sum_{i=0}^{N-1}\left|P_{i} \cap L_{i}\right|+\sum_{i=0}^{N-1}\left|L_{i}-P_{i}\right|+\sum_{i=0}^{N-1}\left|P_{i}-L_{i}\right|} 
    \end{aligned}
\end{equation}

We use in this benchmark the Micro set of metrics (Eq.~\ref{eq:MicroP},~\ref{eq:MicroR},~\ref{eq:MicroF1}) for optimisation, hyperparameter tuning and reporting the results on the test set in Table~\ref{tab:metrics}.

\subsection{Benchmark implementation}
\label{sec:distributed}

We provide a framework, built upon and extending the implementation in \cite{sumbul2020bigearthnet}, for efficient distributed training in TensorFlow API2~\citep{199317}. 
We use Horovod~\citep{sergeev2018horovod}, which is a high-level API that sits on top of TensorFlow for training on multiple nodes and GPUs. In~\cite{sergeev2018horovod}, it is argued that Horovod optimally utilizes the available network to take full advantage of hardware  resources, therefore one could gain better performance than using a pure Distributed TensorFlow implementation. 
Hence, we publish on our github repository the distributed implementation of the deep neural networks pipelines created for this work, along with the pretrained weights to facilitate uptake of novel transfer learning applications based on Sentinel-2 data. 

We conducted our experiments on Aris High Performance Computing (HPC) infrastructure, and used 10 nodes with 2 GPU-NVIDIA Tesla K40 each. In our experiments, we focus on two directions: speed and classification accuracy. We examine how fast we can train quality classifiers, how fast we can classify new samples at inference, and finally, how well our classifiers perform. The metrics systematically recorded are i) training time (hours and minutes), ii) inference rate (images per second), iii) Precision (Eq.~\ref{eq:MicroP}), iv) Recall (Eq.~\ref{eq:MicroR}), v) Accuracy, and vi) F-Score (Eq.~\ref{eq:MicroF1}).


\subsection{Results}


\begin{table*}[h!]
 \begin{tabular}{c c c c c c c c} 
 \toprule
 \multirow{2}{*}{\parbox{0.8 cm}{\textbf{Model}}} & \textbf{Accuracy} & \textbf{Precision} & \textbf{Recall} & \textbf{F-Score} &  \textbf{Training Time} & \textbf{Inference Rate} &  \textbf{Model Size}\\
    &\textbf{(\%)} &\textbf{(\%)} & \textbf{(\%)} & \textbf{(\%)} & \textbf{(hours.mins)} & \textbf{(imgs/sec)} \\
    
 \midrule
 ResNet50$^{1 GPU}$ & 61.8 & 78.1 & 74.8 & \textbf{76.4} & 13.22 & 345 & 23,648,595\\
 \hhline{|=|=|=|=|=|=|=|=}
 ResNet50 & 62.4& 78.8 & 75.0 & 76.8& 0.59 &351& 23,648,595\\
 \hline
 ResNet50-GHOST & 58.4 & 75.2 & 72.4 & 73.8 & 1.04 & 403 & 11,930,387\\
 \hline
 ResNet101 & 61.7& 78.8& 74.0&76.3 &1.38 &268 & 42,719,059\\
 \hline
 ResNet152 & 60.8 & 76.2 & 75.0 & 75.6 & 2.19 & 203& 58,431,827\\
 \hline
 DenseNet121 & 62.3 & 79.2& 74.5 & 76.8 & 0.50 &  \textbf{407} & \textbf{7,078,931}\\
 \hline
 DenseNet169 & 61.1& 77.7&74.1 &75.9 & 1.01&370 & 12,696,467\\
 \hline
 DenseNet201 & 60.8 & 78.1& 73.3& 75.6& 1.17&325 &18,380,435\\
 \hline
 VGG16 & 63& 81.4& 73.6& 77.3&\textbf{0.49} & 242&14,728,467\\
 \hline 
 VGG19 & 63.6&81.7 &74.2 & \textbf{77.7}& 0.57 &218 &20,038,163 \\
 \hhline{|=|=|=|=|=|=|=|=}
K-Branch & 47.3 & 63.4 & 65.1 & 64.2&1.01&510& 36,979,027 \\
 \hhline{|=|=|=|=|=|=|=|=}
 MLPMixer/6 & 57.8 & 76.1 & 70.6 & 73.2 & 0.17 & 654 & 468,083\\
 \hline
 MLPMixer &60.3&79.1&71.8&\textbf{75.2}&0.12& 807& 446,723\\
 \hline
 MLPMixer/20 & 58 & 78.1 & 69.4 & 73.4 & 0.10 & 780 & 740,355\\
\hline
MLPMixer/30 & 53.5 & 77.1 & 63.6 & 69.7& 0.11 & 845 & 1,369,715\\
\hline
MLPMixer/40 &  49.6 & 75.1 & 59.4 & 66.3 & 0.12 & 823 & 2,261,991\\
 \hline
 MLPMixerTiny & 55.7 & 77.5&66.5 &71.6 & \textbf{0.08}& \textbf{911} &  \textbf{40,863}\\
 \hhline{|=|=|=|=|=|=|=|=}
 ViT/6 &54.1 &76.8 &64.6 &70.2 & 2.33 & 241 &55,237,395\\\hline
 ViT/12 & 61.3 &80.7 &71.9 &76 &0.36 &668 & 15,984,915\\ \hline
 ViT/20 & 62.1 & 80.5 & 73.1 & 76.6 &0.23 & \textbf{720} & 7,760,147 \\ \hline
 ViT/30 & 61.6 & 80.1 & 72.7 & 76.2 & 0.20 & 704  &  5,458,707\\ \hline
 ViT/40 & 60.4 & 80.1 & 71 & 75.3 &\textbf{ 0.19} & 691 & \textbf{4,989,203} \\ \hline
 ViTM/20 & 62.7 & 80.6 & 73.8 & \textbf{77.1}& 0.29 &586 & 9,286,419 \\ \hline
 \hhline{|=|=|=|=|=|=|=|=}
  EfficientNetB0-SE & 61.4 & 79.6 & 72.8 & 76.1 & 0.15 & 640 & 4,411,251\\
 \hline
  EfficientNetB0-SE-GHOST & 60.7 & 80.2 & 71.4 & 75.5 & 0.16 & 602 & 3,053,251\\
 \hline
  EfficientNetB0-CBAM & 61.5 & 79.9 & 72.7 & 76.1 & 0.17 & 501 & 4,412,819\\
 \hline
  EfficientNetB0-CBAM-GHOST & 61.0 & 80.5 & 71.7 & 75.8 & 0.18 & 471 & 3,054,819\\
 \hline
  EfficientNetB0-COORD & 61.2 & 79.3 & 72.8 & 75.9 & 0.18 & 604 & 4,191,967\\
 \hline
  EfficientNetB0-COORD-GHOST & 61.3 & 80.0 & 72.4 & 76.0 & 0.19 & 509 & 2,833,967\\
 \hline
  \textbf{EfficientNetB0-ECA} & 61.4 & 79.6 & 72.9 & \textbf{76.1} & \textbf{0.14} & \textbf{651} & \textbf{3,461,663}\\
 \hline
  EfficientNetB0-ECA-GHOST & 61.4 & 80.4 & 72.1 & 76.0 & 0.15 & 611 & 2,103,663\\
 \hhline{|=|=|=|=|=|=|=|=}
  WRNB0 & 56.5 & 80.0 & 65.9 & 72.2 & 0.10 & 807 & 306,803\\
 \hline
  WRNB0-GHOST & 54.9 & 79.5 & 64.0 & 70.9 & 0.11 & 845 & 157,619\\
 \hline
  WRNB0-SE & 61.5 & 81.1 & 71.8 & 76.2 & 0.11 & 751 & 309,729\\
 \hline
  WRNB0-SE-GHOST & 59.6 & 80.8 & 69.5 & 74.7 & 0.12 & 808 & 160,545\\
 \hline
  WRNB0-CBAM & 60.6 & 80.2 & 71.3 & 75.5 & 0.13 & 639 & 310,023\\
 \hline
  WRNB0-CBAM-GHOST & 59.9 & 80.0 & 70.5 & 74.9 & 0.12 & 670 & 160,839\\
 \hline
  WRNB0-COORD & 59.8 & 80.9 & 69.6 & 74.8 & 0.18 & 588 & 312,747\\
 \hline
  WRNB0-COORD-GHOST & 58.3 & 80.4 & 68.0 & 73.7 & 0.19 & 655 & 163,563\\
 \hline
  \textbf{WRNB0-ECA} & 61.7 & 81.5 & 71.8 & \textbf{76.3} & \textbf{0.11} & \textbf{823} & \textbf{306,817}\\
 \hline
  WRNB0-ECA-GHOST & 59.7 & 80.7 & 69.7 & 74.8 & 0.12 & 851 & 157,633\\
 \hline

 \bottomrule \\
\end{tabular}
\caption{Results of the benchmark, conducted with the distributed learning framework, for the eight CNN, two MLP, six transformer, eight EfficientNet-based, and ten WRN-based models. ResNet50$^{1 GPU}$ metrics correspond to the ResNet50 model when trained in one GPU only. For each model family, we highlight in bold the best metric achieved, concerning F-Score, training time, and inference rate. EfficientNetB0-ECA and WRNB0-EC, in bold, are the two best performing models in the benchmark, which we select to scale through compound scaling (Section~\ref{sec:scaleup}).}
\label{tab:metrics}
\end{table*}

\begin{figure*}[ht]

 \includegraphics[width=2\columnwidth]{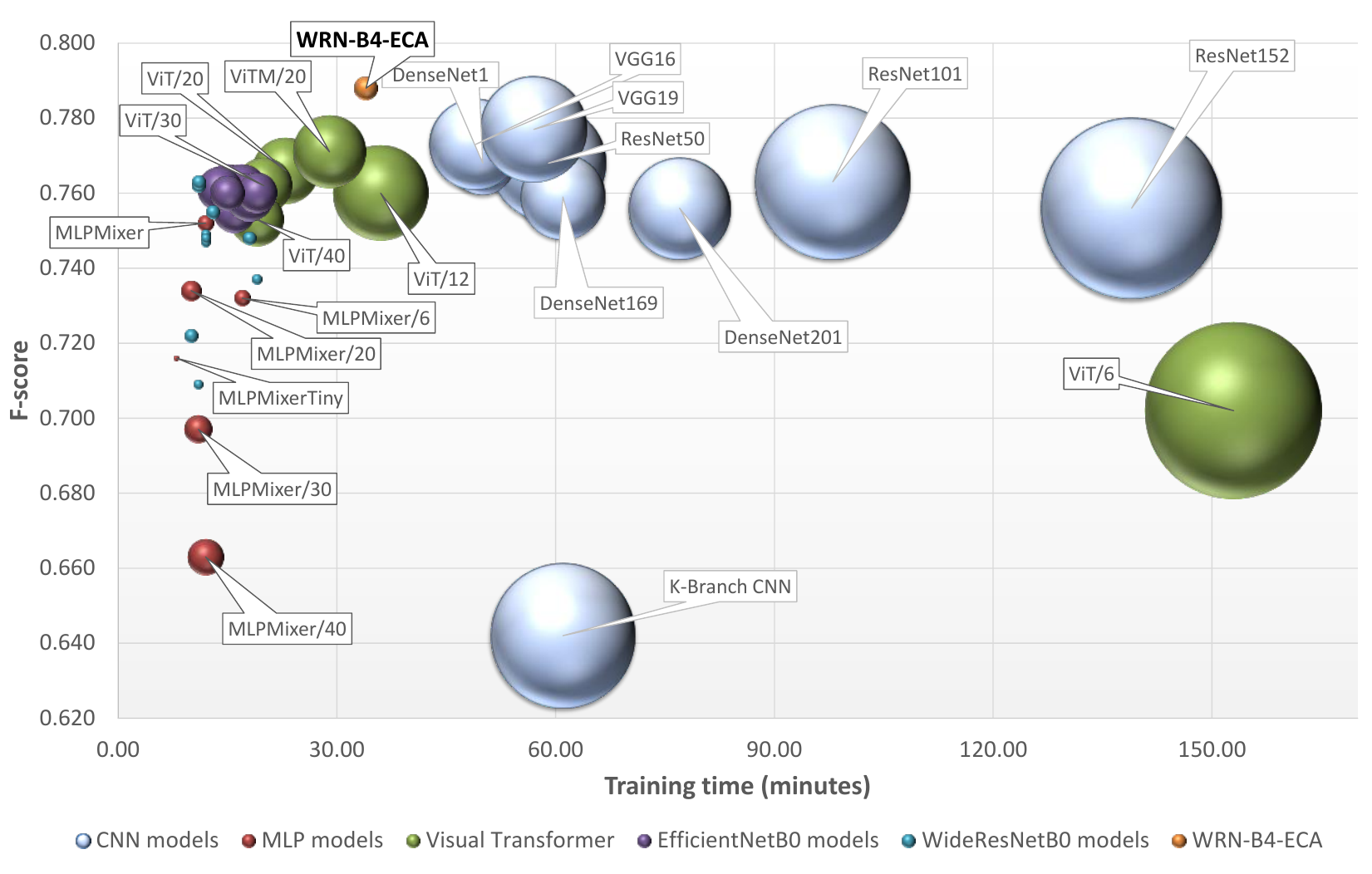}
\caption{Performance vs training time trade-off for the different models evaluated. All models of Table~\ref{tab:metrics} are included, plus our best performing model of Table~\ref{tab:best_model}: WRN-B4-ECA. The size of the bubbles is proportional to the size of the model. Training time is estimated on a cluster of 20 Tesla K40 GPUs.}
\label{fig:performance}
\end{figure*}

We report the benchmark results in Table~\ref{tab:metrics} for the eight CNN, two MLP, six ViT, eight EfficientNet-based and ten WRN-based models. 
In addition, we show in Figure~\ref{fig:performance} the models' performance as a function of the F-score metric (Eq.~\ref{eq:MicroF1}) to capture the classification accuracy, the training time to capture the efficiency of the network, and the total number of trainable parameters. For completeness, we include in Table~\ref{tab:BigEarthNetOrig} the metrics from the original BigEarthNet paper~\citep{sumbul2020bigearthnet} for the CNN architectures tested. 

According to Table~\ref{tab:metrics}, the two best performing models are EfficientNetB0-ECA and WRNB0-ECA. Together with a vanilla EfficientNetB0 model as it is available from Keras, we scale them from B0 to B7 using the compound scaling method (Section~\ref{sec:scaleup}). 
The main difference between our EfficientNet-SE implementation and the EfficientNet-Keras one is the use of the reduction ratio (r) within the Squeeze and Excitation module. We use the reduction ratio suggested by the SE paper authors in~\cite{Hu_2018_CVPR}, instead of the one suggested by the original EfficientNet paper authors~\citep{pmlr-v97-tan19a}.
We present the model scaling results in Table~\ref{tab:best_model}.

\begin{table}[h]
    \centering
    \begin{tabular}{ c c c c}
    \toprule
        Model &  Precision & Recall & F-Score \\
        \hline
        ResNet50 & 81.39 & 77.44 & \textbf{77.11} \\
        \hline
        ResNet101 & 80.18 & 77.45 & 76.49\\
        \hline
        ResNet152 & 81.72& 76.24&76.53\\
        \hline
        VGG16 & 81.05 & 75.85 & 76.01 \\
        \hline
        VGG19 & 79.87 & 76.71 & 75.96\\
        \hline
        \bottomrule\\
    \end{tabular}
    \caption{Model metrics reported by the original BigEarthNet paper~\citep{sumbul2020bigearthnet}. Metrics match relatively well with our implementations presented in Table~\ref{tab:metrics}. Our VGG based model results are slightly better and our ResNet-based model results are slightly worse.} 
    \label{tab:BigEarthNetOrig}
\end{table}


\begin{table*}[]
    \centering
    \begin{tabular}{p{2.5cm}ccccc}
    \toprule
     {\textbf{Model size}} & \textbf{Model } & \textbf{Precision (\%)} & \textbf{Recall (\%)} & \textbf{F-Score (\%)} & \textbf{Training Time (h.mm)} \\
     
        \hline
        \multirow{3}{*}{} WRN-B0-ECA & 306,817 & 81.5 & 71.8 & 76.3 & 0.11 \\
        EfNetB0-ECA & 3,461,663 & 79.6 & 72.9 & 76.1 & 0.14 \\
        EfNetB0-Keras & 4,075,940 & 75.6 &70.9 & 73.1& 0.34 \\
        \hline
       \multirow{3}{*}{} WRN-B1-ECA & 373,381 & 81.8 & 72.5 & 76.9 & 0.12 \\
       EfNetB1-ECA & 5,511,623 & 81.1 & 73.2 & 77.0 & 0.19 \\
       EfNetB1-Keras & 6,601,608 & 75.9 &71.4 & 73.6& 0.49 \\
        \hline
        \multirow{3}{*}{} WRN-B2-ECA & 433,209 & 82.2 & 73.1 & 77.4 & 0.19 \\
        EfNetB2-ECA & 6,503,649 & 81.3 & 73.8 & 77.3 & 0.27 \\
        EfNetB2-Keras & 7,797,370 & 75.0 &70.6 & 72.7 & 0.51 \\
        \hline
        \multirow{3}{*}{}WRN-B3-ECA & 590,333 & 82.4 & 74.4 & 78.2 & 0.23 \\
        EfNetB3-ECA & 8,981,821 & 81.7 & 74.0 & 77.7 & 0.36 \\
        EfNetB3-Keras & 10,815,272  & 76.4& 71.7 & 74.0& 1.03 \\
        \hline
        \multirow{3}{*}{}\textbf{WRN-B4-ECA} & 985,961 & \textbf{82.4} & 75.5 & 78.8 & \textbf{0.34} \\
        EfNetB4-ECA & 14,630,489 & 81.7 & 73.7 & 77.5 & 1.00 \\
        EfNetB4-Keras & 17,710,928 & 73.9 & 72.7 & 73.3 & 1.22 \\
        \hline
        \multirow{3}{*}{}  WRN-B5-ECA & 5,166,299 & 82.0 & \textbf{76.1} & \textbf{79.0} & 2.46 \\
        EfNetB5-ECA & 23,454,139 & 80.6 & 73.9 & 77.1 & 1.37 \\
        EfNetB5-Keras & 28,555,496 & 78.4 & 74.2 & 76.2& 1.55 \\
        \hline
        \multirow{3}{*}{} WRN-B6-ECA & 7,281,895 & 82.1 & 75.1 & 78.5 & 3.50 \\
        EfNetB6-ECA & 33,591,965 & 81.3 & 74.0 & 77.4 & 2.17 \\
        EfNetB6-Keras & 41,007,480 & 75.0 & 72.2 & 73.6 & 2.26 \\
        \hline
        \multirow{3}{*}{}WRN-B7-ECA & 14,068,791 & 79.6 & 73.5 & 76.4 & 7.50 \\
        EfNetB7-ECA & 52,340,949 & 81.6 & 71.4 & 76.1 & 4.09 \\
        EfNetB7-Keras & 64,150,392 & 78.9 & 73.5 & 76.1& 4.45 \\
        \hline
        \bottomrule\\
    \end{tabular}
    \caption{Scaling and training: i) our best performing model: WRNB0-ECA in this table, ii) our EfficientNetB0-ECA (here EfNetB0-ECA) according to Table~\ref{tab:metrics}, and iii) a vanilla EfficientNetB0-Keras (here EfNetB0-Keras) architecture as available from Keras.}
    \label{tab:best_model}
\end{table*}


\section{Discussion}
\label{sec:discussion}
\subsection{Trade-offs in the benchmark}
Training times vary considerably for the traditional CNN family of models, ranging from $\sim$50 minutes to over two hours, while F-score varies within a 2\% interval. Overall VGG prevails. VGG19 achieves the highest accuracy, trained in 57 minutes, while VGG16 is the fastest to train (49 minutes), even though it has more parameters than some of its CNN counterparts, e.g. DenseNet169. At inference though, the fewer the overall parameters, the higher the processed images per second rate, and here DenseNet121 performs best. The CNN model results of Table~\ref{tab:metrics} match well with the metrics reported in \citet{sumbul2020bigearthnet} (Table~\ref{tab:BigEarthNetOrig}). Furthermore, we train a ResNet50 model on only 1 GPU-NVIDIA Tesla K40, in order to appreciate the improvement in training time with our distributed learning implementation (Section~\ref{sec:distributed}). Rows 1 and 2 in Table~\ref{tab:metrics} highlight the 13.5$\times$ speedup with our implementation. Moreover, we evaluate the impact, the Ghost module has on the ResNet50 model. We notice a 50\% decrease of the model parameters and therefore an improvement inference rate, accompanied by an almost 8\% increase in the training time, as well as a 3\% drop of the F-score, compared to the ResNet50 model. Our results for the ResNet50-GHOST model, match well with the results reported by ~\cite{Han_2020_CVPR}.

The MLPmixer-based models are very efficient in training and inference, but are slightly below average in classification accuracy. In practice, these models can achieve performance similar to the more heavyweight CNN models, with significantly less parameters, from as little as 40 thousand parameters. Large CNN models, such as ResNet152 and DenseNet201, achieve almost the same accuracy as MLPmixer, but use around 58 and 18 million parameters respectively. Such a large model capacity has negative effects to both training time and memory usage. MLPMixer presents a good balance between training time and overall accuracy, with just under half a million trainable parameters. 

On the other hand, our MLPMixerTiny model has the fewest trainable parameters, for example, three orders of magnitude less compared with ResNet101, and is trained for 30 epochs in just eight minutes, which is the fastest time in the benchmark. This is a $6\times$ improvement with respect to the fastest traditional CNN model, VGG16, and a $17\times$ improvement with respect to the the slowest to train CNN model, ResNet152. This comes at the expense of accuracy though: with 71.6\% F-score, MLPMixerTiny is at the bottom of the list, 
 hence in this case there is a trade-off between super-fast training time versus a drop in classification accuracy, as one would expect.

\begin{figure}[ht]
\begin{subfigure}{0.23\textwidth}
\centering
 \includegraphics[width=\textwidth]{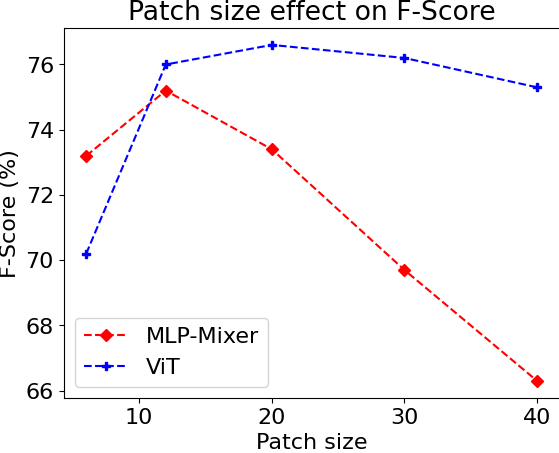}
\end{subfigure}
\begin{subfigure}{0.23\textwidth}
\centering
\includegraphics[width=\textwidth]{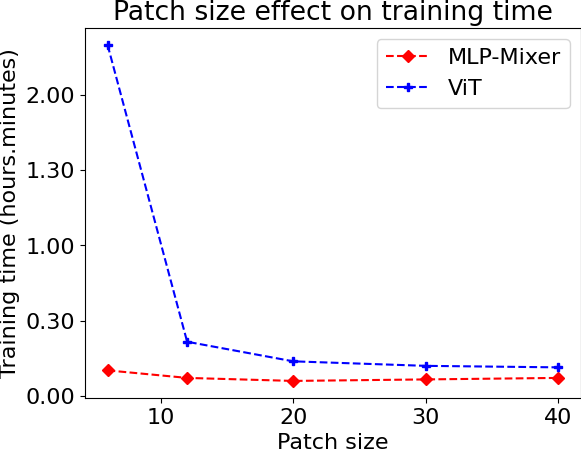}
\end{subfigure}
\caption{Ablation on the effect of patch size on model's performance in regards to F-Score and training time for both MLPMixer and Vision Transformer.}
\label{fig:patch_effect}
\end{figure}

This trade-off does not apply to ViT, EfficientNets and WRNs models, for different reasons. Interestingly, ViT/6 produces the lowest accuracy and is also the most difficult to train. It experiences the worse overall performance (accuracy and training time) in the study. Examining the effect of the patch size, we show that very small values have a negative effect on the performance of the model. Patch size of 20 produces the best results with ViT/20 achieving 77.1\% F-Score, 
 while retaining the training time under half an hour. This is matching the accuracy of the best CNNs, but the model is trained faster. A similar effect is observed for the MLP-Mixer. Our base version with patch size equal to 12 produces the best F-Score while maintaining a good training time. Increasing the patch size too much significantly hurts the performance of MLP-Mixer with no considerable gain in training speed. The effect of patch size is summarized in Figure~\ref{fig:patch_effect} for both architectures vis-à-vis F-Score and training time.
 
EfficientNets and even more WRNs model families, exhibit the best overall performance considering the trade-offs between training time, F-score, and inference rate. This happens even for the non-scaled, simpler B0 models. Both EfficientNet and WRN models share four common characteristics: i) they achieve $4\times$ to $10\times$ faster training times compared to CNNs, ii) the inclusion of the Ghost module deteriorates their performance mainly in terms of classification accuracy, in contrast to the author claims, while reducing the models' size by $\sim$35\% and $\sim$50\% for EfficientNets and WRNs respectively. iii) even though including the Ghost module significantly reduces the models' size, those models require a fraction of additional time to train compared to the models without the Ghost module. This is because the Ghost module implementation is more suitable for ARM/CPUs and is not GPU-friendly due to the Depthwise Convolution operations, and therefore is more appropriate to deploy in mobile devices and other devices with limited resources. For these reasons, we believe that the Ghost versions of our models are a promising candidate for Inference-at-the-Edge use cases. 
Finally, iv) the ECA attention module provides the best overall results for F-score and training time. WRNs in particular perform best (Figure~\ref{fig:performance} and Table~\ref{tab:metrics}), considering the fact that they have one order of magnitude fewer parameters with respect to EfficientNets. This does not come at the expense of classification accuracy and on the contrary, it is directly translated to better inference times. 
Finally, it is noteworthy that classification accuracy variance is negligible for the variants of EfficientNet based models (Figure~\ref{fig:performance}), while this does not apply to WRN variants. The attention mechanism in WRN affects their performance significantly. 

\subsection{The new SOTA for BigEarthNet}
EfficientNetB0-ECA and WRNB0-ECA models (Table~\ref{tab:metrics}) are scaled through compound scaling and the results are reported in Table~\ref{tab:best_model}. According to Table~\ref{tab:best_model}, we select our top model to be WRN-B4-ECA with an F-score of 78.8\% trained in 34 minutes and processing 381.0 images/sec at inference. According to our literature review, this is the new SOTA for the BigEarthNet dataset. Although WRN-B5-ECA reaches an F-score of 79.0\%, the 0.2\% gain is not enough to justify the 5$\times$ training time and model parameters. In fact, we observe that for WRN-B6-ECA and WRN-B7-ECA  models, F-score drops, hinting at overfitting. Additional training data would be needed to estimate the weights for these larger models. 


The best model in terms of classification accuracy, trained in the original BigEarthNet paper~\citep{sumbul2020bigearthnet} is ResNet50 with approximately 23 million parameters. In that work, the authors achieved an image classification F-score of 77.11\% (Table~\ref{tab:BigEarthNetOrig}), while with our setup we reached 76.8\% and trained it in almost one hour (Table~\ref{tab:metrics}). On the other hand, our lighter WRN-B4-ECA model with one million parameters only manages an F-score of 78.8\%, and is trained on the same data splits in 34 minutes, which constitutes a significant improvement. This is visually highlighted in Figure~\ref{fig:performance}. We further investigate this improvement by looking into the metrics for each one of the 19 classes that exist in the dataset. These are shown in Table~\ref{tab:19classes} for the baseline ResNet50 model, vis-à-vis our WRN-B4-ECA model. For every single class, our model outperforms the baseline. For some classes the difference is remarkable, for example, class `permanent crops' is better resolved with an F-score jump from 52\% to 65.6\%. Similar improvement jumps are noted for `transitional woodland-shrub' and `coastal water' classes. Overall, the averaged metrics in the last line of Table~\ref{tab:19classes}, which correspond to Macro F-score instead of the Micro F-score measures in Equation~\ref{eq:MicroF1}, show an increase by almost 4.47\% when using our lighter model.

\subsection{Examination of the effect of input resolution on our SOTA model}
\begin{figure}[ht]
\centering
 \includegraphics[width=0.45\textwidth]{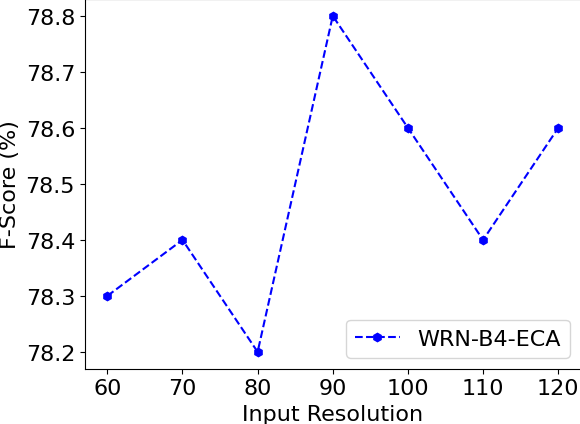}
\caption{The effect of various input resolutions on the performance of our WRN-B4-ECA model.}
\label{fig:wrn_b4_eca_input_res}
\end{figure}

Following the designation of our WRN-B4-ECA model as the new SOTA for BigEarthNet, we investigate whether we could benefit in terms of performance by altering the input image resolution. The default input resolution used by our WRN-B4-ECA model is 90$\times$90, while it ranges during the course of our scaling experiments from 60$\times$60, up to 120$\times$120. The impact of the different input image resolutions are summarized in Figure \ref{fig:wrn_b4_eca_input_res}. Our investigation demonstrates that the default input resolution used by the WRN-B4-ECA model is ideal and results in the highest performance among all the other variants.

\subsection{Examination of the effect of input channels}
\begin{figure}[ht]
\centering
 \includegraphics[width=0.45\textwidth]{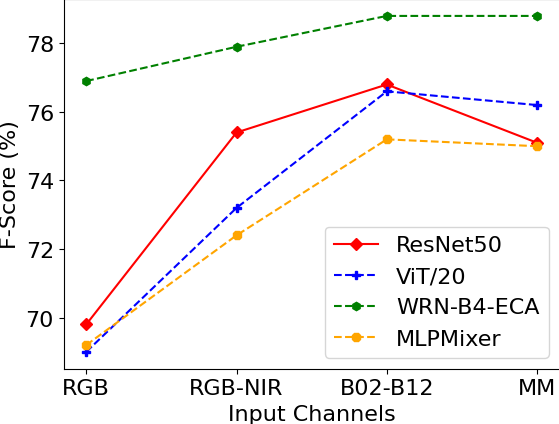}
\caption{Ablation on the effect of input channels. X-axis represents the input setting. RGB refers to models trained solely with the RGB bands as input, RGB-NIR corresponds to models trained with RGB and the NIR channel, B02-B12 is our normal setup and finally MM combines both B02-B12 channels of Sentinel-2 and VV-VH channels of Sentinel-1.}
\label{fig:channel_ablation}
\end{figure}

 Motivated by the introduction of the SAR modality (Sentinel-1) in the second version of BigEarthNet dataset \citep{sumbul2021bigearthnetmm}, we investigate whether the extra information provided by the SAR images and the different multispectral channels of Sentinel-2 are actually beneficial for the task at hand. Our investigation revolves around four diverse architectures, i.e. ViT, ResNet50, MLPMixer and our best performing model WRN-B4-ECA. We define the following settings for our experiments. First, we experiment with inputs consisting solely of the RGB channels. We then augment our input by introducing the NIR band and train with 4-channel inputs. Since we have already conducted the experiments with all multispectral (B02-B12) channels, we proceed with the introduction of the Sentinel-1 modality resulting in a 12-channel input. The results of this ablation are summarized in Figure \ref{fig:channel_ablation}. From our experiments, it is clear that the multispectral information is very beneficial for the task of land use land cover classification. This finding, emphasizes our initial intuition that the restriction on the input channels (RGB) induced by models pretrained on ImageNet results in loss in information.
 
 On the other hand, Sentinel-1 data do not seem to contribute much for this task. This could be attributed to the nature of the dataset. BigEarthNet is focused on frames with minimum cloud coverage. SAR data could prove to be really helpful in scenarios of high cloud coverage, since the radar microwave frequencies can penetrate clouds providing some backscatter information, as opposed to optical data that are completely obscured by clouds.  
 
 Finally, these experiments emphasize the superiority of our proposed WRN-B4-ECA model, as it consistently outperforms the rest of the models and maintains its performance even after the introduction of the Sentinel-1 images contrary to the rest of the models in this experiment.

\begin{table}[]
    \centering
    \begin{tabular}{ c c c }
    \toprule
    \multirow{2}{*}{\textbf{BigEarthNet class}} & \textbf{BigEarthNet} & \textbf{Our} \\
    & \textbf{ResNet50} & \textbf{WRN-B4-ECA} \\
        \hline
       Urban fabric & 74.84 & 75.17 \\
        \hline
        \parbox{3.5 cm}{\centering Industrial or commercial units} & 48.55 & 49.14 \\
        \hline
        Arable land & 83.85 & 86.25 \\
        \hline
        Permanent crops & 51.91 & 65.49 \\
        \hline
        Pastures & 72.38 & 76.27 \\
        \hline
         \parbox{3.5 cm}{\centering Complex cultivation patterns} & 66.03 & 70.27 \\
        \hline
        \parbox{3.5 cm}{\centering Land principally occupied by agriculture, with significant areas of natural vegetation} & 60.94 & 65.89 \\
        \hline
        Agro-forestry areas & 70.49 & 77.89 \\
        \hline
         Broad-leaved forest & 74.05 & 80.48 \\
        \hline
         Coniferous forest & 85.41 & 86.97 \\
        \hline
         Mixed forest & 79.44 & 81.24 \\
        \hline
         \parbox{3.5 cm}{\centering Natural grassland and sparsely vegetated areas} & 47.55 & 51.28 \\
        \hline
        \parbox{3.5 cm}{\centering Moors, heathland and sclerophyllous vegetation} & 59.41 & 62.54 \\
        \hline
         \parbox{3.5 cm}{\centering Transitional woodland-shrub} & 53.47 & 68.1 \\
        \hline
         Beaches, dunes, sands & 61.46 & 66.67 \\
        \hline
         Inland wetlands & 60.64 & 58.47 \\
        \hline
         Coastal wetlands & 47.71 & 63.16 \\
        \hline
         Inland waters & 83.69 & 86.06 \\
        \hline
         Marine waters & 97.53 & 98.57 \\
        \hline
        \textbf{Average} & \textbf{67.33} & \textbf{71.8} \\
        \hline
        \bottomrule\\
    \end{tabular}
    \caption{BigEarthNet class-based F-scores (\%) for our best performing WRN-B4-ECA model according to Table~\ref{tab:best_model}, and a ResNet50 baseline CNN model.}
    \label{tab:19classes}
\end{table}

 \begin{figure*}[!ht]
 \includegraphics[width=1.8\columnwidth]{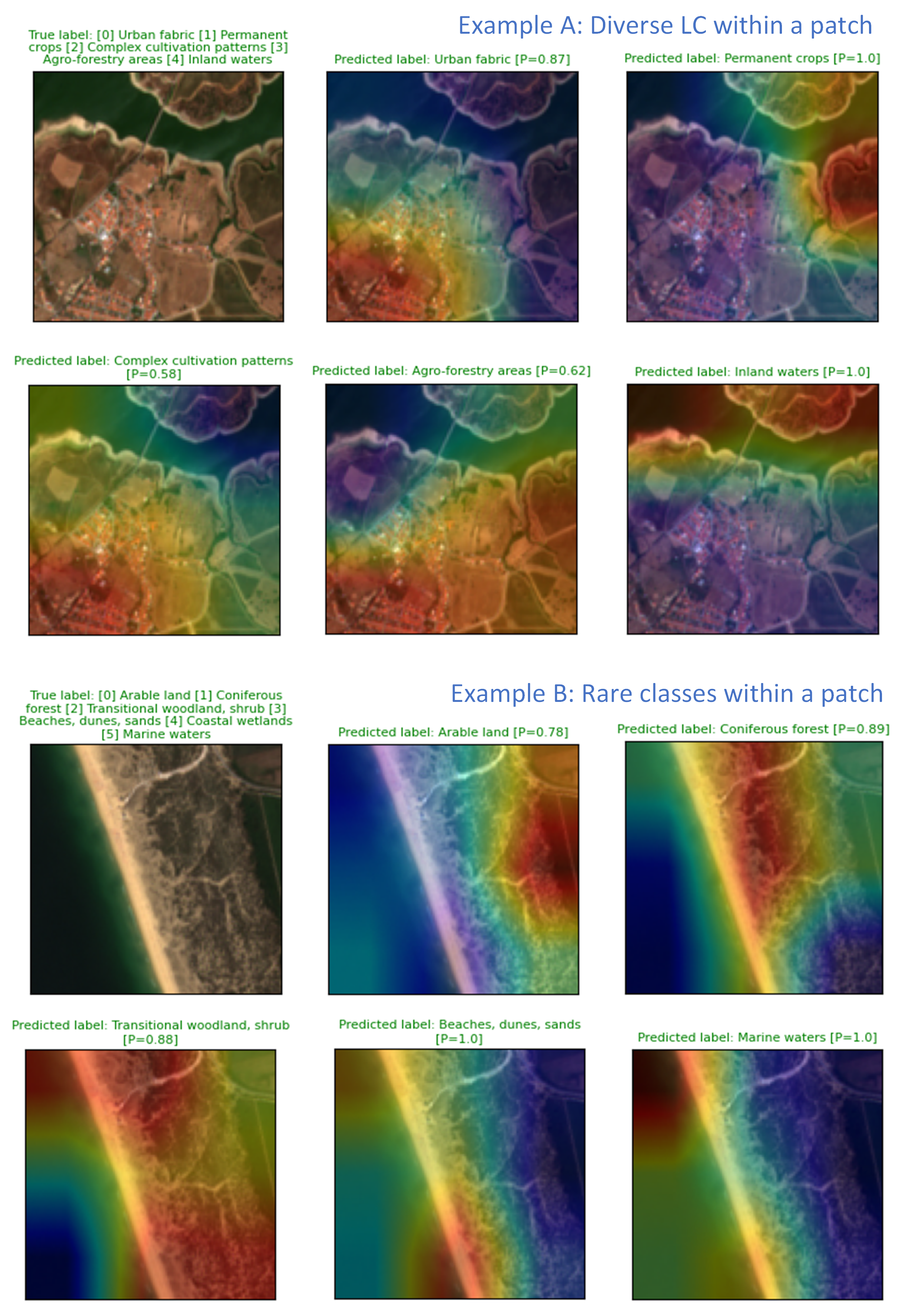}
 \centering
\caption{Examples of two challenging image patches, correctly classified. The first patch in Examples A \& B is the original Sentinel-2 image patch with the different LULC classes contained. The other patches are the output of Grad-CAM~\citep{Selvaraju_2017_ICCV} that we adopt to interpret which parts of the image were used by our network for deciding on each specific True Positive LULC class. P stands for the probability a specific LULC is contained in the patch.}
\label{fig:GradCamTP}
\end{figure*}

\subsection{Model explainability}
We use Gradient-weighted Class Activation Mapping (Grad-CAM), as in~\cite{Selvaraju_2017_ICCV}, in order to understand some of the image classification accuracy discrepancies observed in the benchmark. Grad-CAM produces `interpretable' explanations for the classification decisions of our resulting model. It exploits the gradients of logits for the different classes of the final convolutional layer to produce a map that  highlights the important regions in the image, used for predicting a specific class.

In Figure~\ref{fig:GradCamTP} we have selected two challenging Sentinel-2 image patches, one in each row, that contain several LULC classes. We also show the Grad-CAM output for the True Positive classes predicted, with the associated probability P that a specific LULC is contained in the patch. If P $>$ 0.5 we assign this class to the patch. On the top of Figure~\ref{fig:GradCamTP} is an image patch with urban, agricultural, vegetated and marine areas, and all different classes are correctly resolved, while the classification decision seems to focus on the appropriate parts of the image. On the bottom of Figure~\ref{fig:GradCamTP} is an image patch with some rare classes, e.g. \textit{Beaches, dunes, sands} and \textit{Transitional woodland, shrub}, which are predicted with high probabilities, and again focusing on the correct part of each image.

 \begin{figure*}[h]
 \includegraphics[width=2\columnwidth]{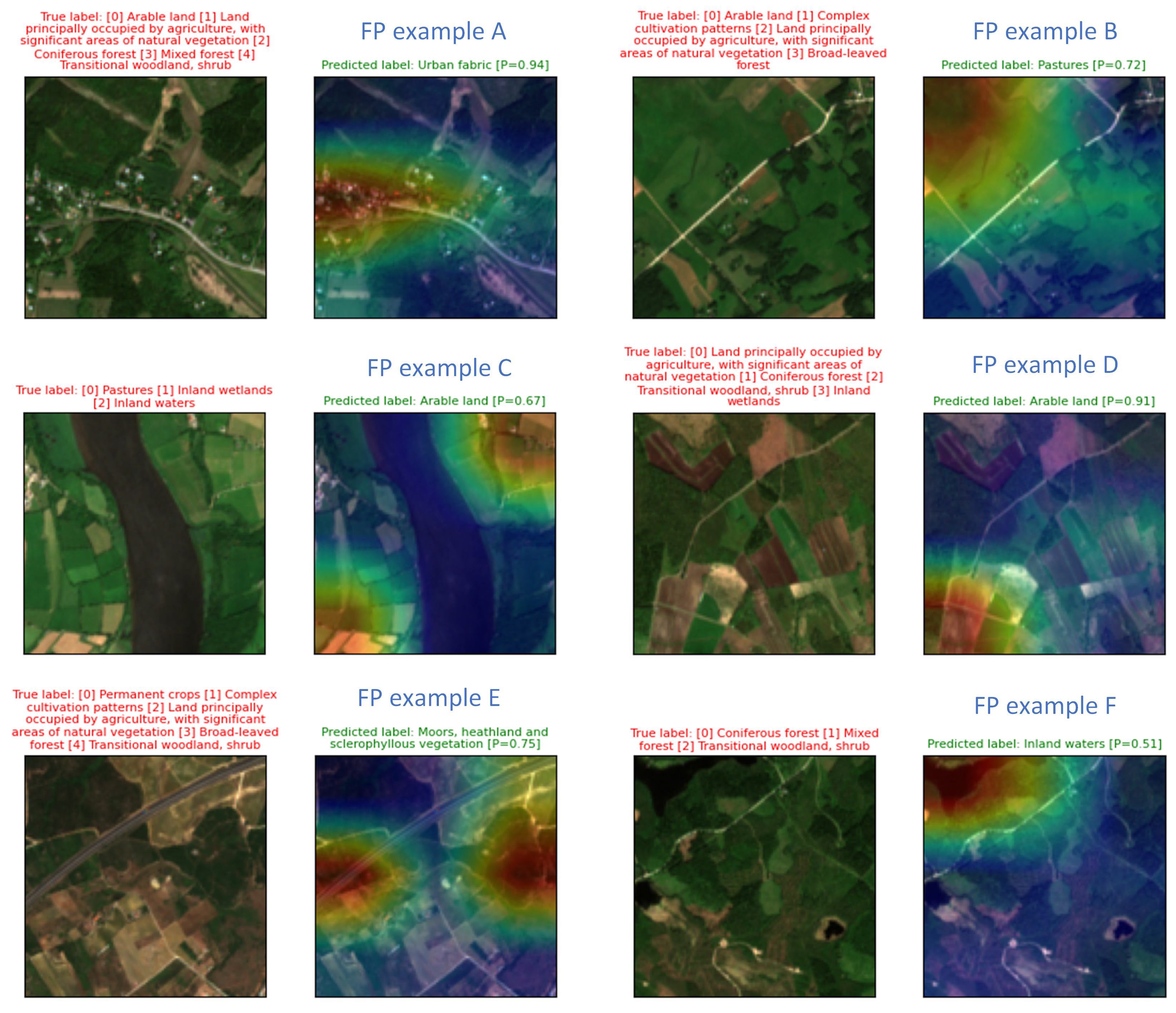}
\caption{Examples of image pairs with False Positive LULC {\color{blue} scene} classification. For each pair we show on the left the original Sentinel-2 image patch with all the LULC classes contained, and we show on the right the Grad-CAM~\citep{Selvaraju_2017_ICCV} output for a False Positive class. Red areas correspond to the part of the image patch used to make the False Positive prediction. P stands for the probability a specific LULC is contained in the patch.  }
\label{fig:GradCamFP}
\end{figure*}

In Figure~\ref{fig:GradCamFP} we provide False Positive samples and the corresponding Grad-CAM output. Investigating these Grad-CAM outputs and relying on the visual content of the visible spectral channels only, it can be argued that it is challenging for the human eye to reject the predicted False Positive LULC classes as well. For example, the \textit{Urban fabric} predicted class in Figure~\ref{fig:GradCamFP}, indeed focuses on settlements or individual buildings that exist in the original image patch. Similarly, the \textit{Pastures} predicted class cannot be easily dismissed, especially when not all image spectral content is visualized. FP class \textit{Arable land} is attributed to patches that according to Grad-CAM output indeed focus on agricultural-like areas. In this case, the higher level category is correct, i.e. agricultural areas, while the more detailed LULC class in the taxonomy is not correct, i.e. \textit{Arable land} is predicted instead of \textit{Land principally occupied by agriculture, with significant areas of natural vegetation}. The last sample in Figure~\ref{fig:GradCamFP} contains an \textit{Inland waters} FP class prediction, and the network focuses on the upper left part of the patch which indeed could be attributed to a water body. Overall, it could be argued that the FP predictions are within the error margin even of an experienced remote sensing photo-interpreter. 

\begin{figure*}[!ht]
\centering
 \includegraphics[width=2\columnwidth]{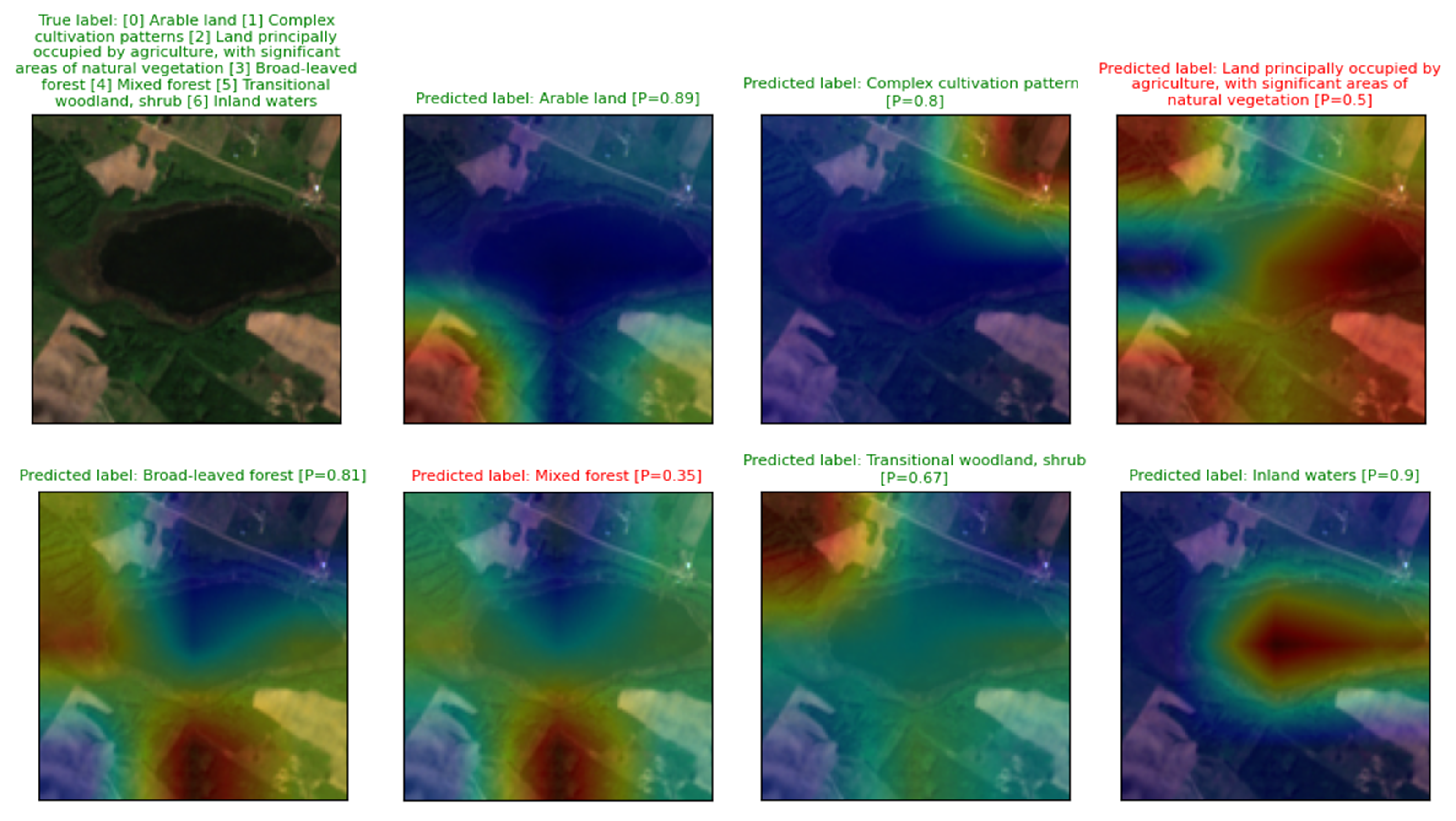}
\caption{Example of an image patch with True Positive and False Negative LULC scene classification. The top-right image is the original Sentinel-2 image patch with all the LULC classes contained. The other images are the output of Grad-CAM~\citep{Selvaraju_2017_ICCV} for interpreting which parts of the original patch were used by our network for deciding on each specific True Positive (green font) or False Negative (red font) LULC class.P stands for the probability a specific LULC is contained in the patch.}
\label{fig:GradCamFN}
\end{figure*}

Finally, in Figure~\ref{fig:GradCamFN} we show an example of an image patch that contains seven different LULC classes, of which five are correctly predicted and two are False Negatives. Grad-CAM outputs for the TP classes again focus visually on the correct parts of the patch. The \textit{Inland waters} class especially has a Grad-CAM output that could be potentially used directly for image segmentation. Investigating hundreds Grad-CAM samples in our test dataset, the same can be inferred for several other LULC predictions. The FN LULC classes are hard to predict, even for an expert in satellite image photointerpretation. Indicatively, while Grad-CAM focuses on densely vegetated areas in Figure~\ref{fig:GradCamFN}, our network correctly predicts the \textit{Broad-leaved forest} class but not the \textit{Mixed forest} class. These two classes, however, are almost indistinguishable considering their spectral signatures. Therefore it could be worth investigating creating a new taxonomy by merging of LULC classes, for which the spectral content and spatial patterns are so similar that even deep neural networks cannot confidently resolve.


\subsection{Transfer learning}

BigEarthNet revolves around a specific RS task, LULC scene classification with a specific set of classes. In other RS tasks that use Sentinel-2 imagery, researchers are developing new training datasets on an ad-hoc basis. Creating an independent, unique, labeled dataset for each RS problem is not feasible, given the plethora of tasks in the RS domain. The paradigm in the Computer Vision community is different. The publicly available models trained on ImageNet natural images have been successfully and extensively used in various transfer learning applications. Motivated by the impact of this approach, we believe that a similar logic should be adopted by the remote sensing community. However, transferring knowledge from such a different domain is not optimal. Having this in mind, we provide a Sentinel-2 domain-relevant pre-trained model zoo, which can be subsequently used to the fullest for different transfer learning RS applications.

We put this argument to the test, with an extensive study on the performance of our models pretrained on BigEarthNet for new RS tasks. We split our investigation in two experiments on two different datasets: EUROSAT~\citep{helber2019eurosat} and SEN12MS~\citep{Schmitt2021}, introduced in Section~\ref{sec:otherdata}. EUROSAT is a Sentinel-2 based dataset addressing the task of land cover classification. 
Given that there is no standard dataset split, we divide EUROSAT in three sets for training, validation and testing with a ratio of 80/10/10. SEN12MS is the second Sentinel based dataset we use for this study, which contains both Sentinel-1 and Sentinel-2 modalities.
With SEN12MS, we evaluate our models on a multi-label scene classification problem, and it is a more challenging dataset compared to EUROSAT. The current state of the art reported in \citep{Schmitt2021} achieves at most 69.9\% F-Score, when using only the Sentinel-2 modality as input to a ResNet50 and 72.0\% with a DenseNet121 when combining both Sentinel-1 and Sentinel-2. Furthermore, the authors notice the importance of the multispectral information, observing a drop in performance when ignoring it. On the other hand, the authors of \citep{helber2019eurosat} achieve very high accuracy on EUROSAT, i.e. 98.56\% using solely the RGB channels as input to a ResNet50 model.
Moreover, their experiments show that one can achieve very high accuracy $>90\%$ while using only one band making the multispectral information redundant. A reason for this could be the choice of the classes in EUROSAT, consisting mainly of high level categories that could be discriminated by the shapes, texture and color e.g. Sea\&Lake versus Highway. On the contrary, SEN12MS, similar to BigEarthNet, aims to identify land use and land cover in a more fine-grained manner, containing classes such as Grassland, Cropland and Shrubland.

In our experiments, we investigate a) the quality of representations learnt on BigEarthNet compared to the respective representations of ImageNet, b) the benefit of the introduction of pretrained models that can exploit the full multispectral information, and c) the capacity of these models for good performance in low data regimes. Overall, we observe a consistent improvement induced by the usage of weights learnt on BigEarthNet.

\begin{figure}[!htp]
\begin{subfigure}{0.23\textwidth}

\centering
 \includegraphics[width=\columnwidth]{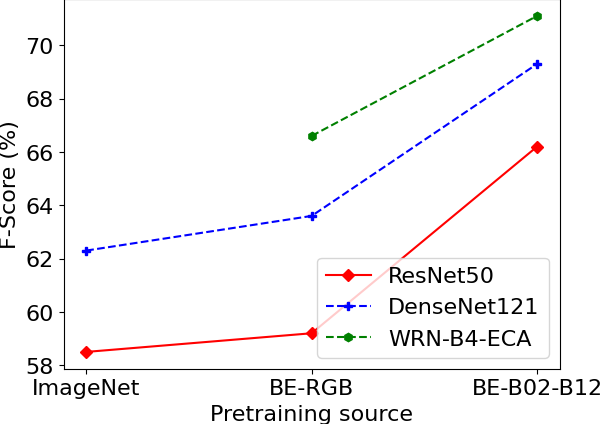}
 \caption{}
 \label{fig:fullytrainablesen12ms}
 \end{subfigure}
 \begin{subfigure}{0.23\textwidth}
 \includegraphics[width=\columnwidth]{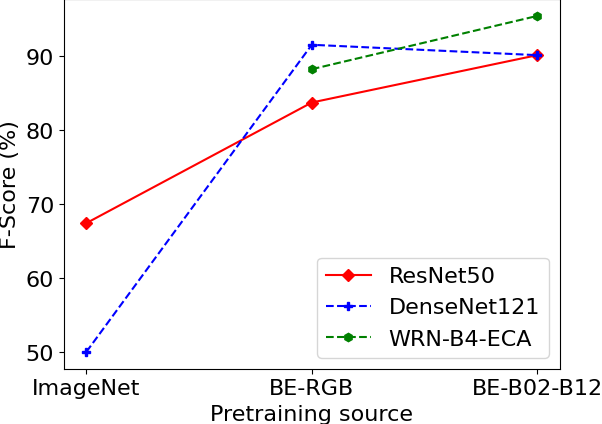}
 \caption{}
 \label{fig:fullytrainableeurosat}
 \end{subfigure}
\caption{Finetuning of ResNet50, DenseNet121 and WRN-B4-ECA under different pretraining schemes. We present the results of the evaluation on SEN12MS and EUROSAT in a) and b) respectively.}
\label{fig:finetuningsource}
\end{figure}

We begin our study by comparing the models pretrained on ImageNet and the same architectures pretrained on BigEarthNet to prove the superiority of in-domain learnt features. We use common architectures with existing ImageNet pretrained weights, i.e. ResNet50 and DenseNet121. Additionally, we include our top performing model WRN-B4-ECA to examine its performance on different datasets. Since our derived model is introduced in this work there are no weights pretrained on ImageNet. Given that ImageNet contains only RGB images, we include in our study the respective setting pretrained on BigEarthNet. For our first experiment then, summarized in Figure \ref{fig:finetuningsource}, we finetune our pretrained models on SEN12MS and EUROSAT examining three pretraining schemes: pretraining on ImageNet, pretraining on BigEarthNet using solely the RGB channels as input, and pretraining on BigEarthNet using the full multispectral information as done in Table \ref{tab:metrics}. We allow all layers to be trainable in the finetuning phase. The superiority of the models pretrained on BigEarthNet is a fact for all experiments (Figure \ref{fig:finetuningsource}), even when the models were trained solely with RGB inputs. All models benefit by the transition from ImageNet to BigEarthNet pretraining scheme.
 WRN-B4-ECA seems to consistently perform the best, achieving the best results when fed with multispectral inputs. This information could not be harnessed by models pretrained on ImageNet without any architecture modification. Nevertheless, these results are achieved after training on full, curated datasets. To evaluate the impact of in-domain transfer learning for RS scenarios where creating a new large annotated dataset is not feasible, we have to investigate the performance of our models in different data-regimes.

\begin{figure}[!htp]
\centering
 \includegraphics[width=\columnwidth]{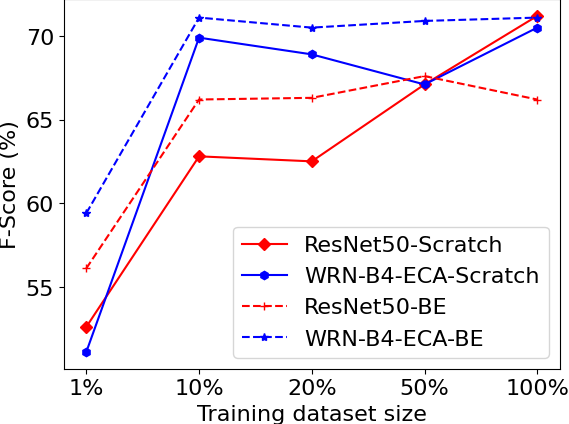}
\caption{ Investigation of transfer learning on low data regimes on SEN12MS. We indicate the ResNet50 with red and the WRN-B4-ECA with blue. The dashed lines show the performance of the models pretrained on BigEarthNet(BE) while the solid lines the performance of models trained from scratch.}
\label{fig:lowdataregime}
\end{figure}

Hence, in our second experiment, we follow this intuition and investigate the behaviour of our models in lower data regimes. For this examination we focus on ResNet50 and WRN-B4-ECA trained on variants of SEN12MS with the multispectral channels as input. SEN12MS's volume allows us to examine a wider range of dataset sizes. We attempt to learn the classification task defined in SEN12MS using the 1\%, 10\%, 20\%, 50\% and 100\% of the training dataset and observe the difference in performance when compared to random initialization. In both initialization settings, we train the whole network. The results of this experiment are shown in \ref{fig:lowdataregime}. As expected, pretraining on large curated datasets such as BigEarthNet can significantly improve performance for related data. Both models pretrained on BigEarthNet perform consistently better than their randomly initialized counterparts. When training with very small training datasets, the performance boost induced by pretraining is very high for both architectures. Impressively, WRN-B4-ECA is able to achieve state-of-the art performance ($71.1\%$ F-Score) with just 10\% of the dataset. Naturally, as the dataset size increases the need for pretrained weights is limited and the difference between random initialization and pretraining diminishes.

These findings highlight the fact that the pretrained model-zoo can be beneficial for RS tasks with small labeled sets, as well as for research labs with low computational resources.
Furthermore, by providing a large enough and reproducible benchmark, the evaluation of future methods becomes more transparent. Finally, the usage of pretrained models as well as a common reference benchmark alleviates the need for repeating expensive experiments leading to reduced carbon footprint. 

\section{Conclusion}
We address the multi-label, multi-class LULC single Sentinel-2 image classification problem and benchmark several popular deep learning architectures and more sophisticated models, such as ViT. We develop and use a distributed learning implementation, and create a model zoo of 60 trained models, which we make publicly available in order to boost research in diverse tasks that exploit multispectral imagery. Considering the challenges in training on big satellite datasets, we seek to optimize model performance jointly in terms of training time, inference rate and classification accuracy. We find that through compound scaling of lightweight Wide Residual Networks we achieve the best overall performance. Our lightweight Wide Residual Network model with an Efficient Channel Attention mechanism and scaled by adapting the EfficientNet compound scaling methodology performs best in our benchmark. It achieves 4.5\% higher averaged f-score classification accuracy for all 19 LULC classes or a 3\% increase in overall micro f-score, and is trained two times faster compared to a ResNet50 baseline model. Finally, our benchmark reveals that conventional CNN models that perform costly convolutions can be matched by Multilayer Perceptron feedforward artificial neural networks that are more lightweight, much simpler, and faster to train. 


Our findings imply that efficient lightweight deep learning models that are fast to train when appropriately scaled for depth, width, and input data resolution can provide comparable and even higher image classification accuracies. This is especially important in remote sensing where the volume of data coming from the Sentinel family but also other satellite platforms is very large and constantly increasing. We believe that our approach for designing light and scalable models can go beyond the specific LULC scene classification problem addressed herein, and could be tested in different application scenarios, e.g. in food security at large scales, and other tasks, such as semantic segmentation and object detection in satellite imagery. 
However, the potential for transfer learning of the DL models has to be thoroughly investigated, especially considering the spatio-temporal nature of satellite data. As discussed by~\cite{sykas2022sentinel}, the classification generalisation capacity to different years and geographic locations remains a great challenge in RS, and domain adaptation strategies should be adopted to bridge the inherent gaps.

We hope that this extended benchmark will serve as a quick and robust way for the evaluation of new methods, and the produced model zoo will propel deep learning research in currently, untouched, applications of remote sensing.

\section*{Acknowledgment}
 This work has received funding from the European Union’s Horizon2020 research and innovation project DeepCube, under grant agreement number 101004188. In addition, this work was supported by computational time granted from the National Infrastructures for Research and Technology S.A. (GRNET S.A.) in the National HPC facility - ARIS - under project ID pr010006.
\bibliographystyle{bibstyles/model1-num-names}
\bibliography{arxivbibliography.bib}

\end{document}